\documentclass[final]{cvpr}

\usepackage{times}
\usepackage{epsfig}
\usepackage{graphicx}
\usepackage{amsmath}
\usepackage{amssymb}
\usepackage{booktabs}
\usepackage{multirow}
\usepackage{hhline}
\usepackage{bbding}
\usepackage{color}

\usepackage{pifont}

\usepackage[pagebackref=true,breaklinks=true,colorlinks,bookmarks=false]{hyperref}

\begin{document}

\title{Contrastive Attention Network with Dense Field Estimation for Face Completion}



\author{Xin Ma${^{1,2}\dag}$, Xiaoqiang Zhou${^{2,4}\dag}$, Huaibo Huang${^{1,2}*}$, Gengyun Jia$^{1,2}$, Zhenhua Chai$^{2}$, Xiaolin Wei$^{2}$ \\
{$^1$}School of Artificial Intelligence, University of Chinese Academy of Sciences\\
{$^2$}NLPR$\;\&\;$CEBSIT$\;\&\;$CRIPAC, CASIA \\
{$^3$}Vision Intelligence Department, Meituan \\
{$^4$}University of Science and Technology of China\\
{\tt\small \{xin.ma, huaibo.huang, gengyun.jia\}@cripac.ia.ac.cn, xq525@mail.ustc.edu.cn}\\
{\tt\small \{chaizhenhua, weixiaolin02\}@meituan.com}
}

\maketitle
\renewcommand{\thefootnote}{\fnsymbol{footnote}}
\footnotetext{* indicates the correspondence author}
\footnotetext{$\dag$ Xin Ma and Xiaoqiang Zhou have contributed equally to the work}
\footnotetext{\href{https://www.sciencedirect.com/science/article/abs/pii/S0031320321006415}{The link to the Pattern Recgnition version}}
\footnotetext{\href{https://github.com/XinMa-AI/CANDE2FC}{The link to the codes}}

\begin{abstract}
Most modern face completion approaches adopt an autoencoder or its variants to restore missing regions in face images. Encoders are often utilized to learn powerful representations that play an important role in meeting the challenges of sophisticated learning tasks. Specifically, various kinds of masks are often presented in face images in the wild, forming complex patterns, especially in this hard period of COVID-19. It's difficult for encoders to capture such powerful representations under this complex situation. To address this challenge, we propose a self-supervised Siamese inference network to improve the generalization and robustness of encoders. It can encode contextual semantics from full-resolution images and obtain more discriminative representations. To deal with geometric variations of face images, a dense correspondence field is integrated into the network. We further propose a multi-scale decoder with a novel dual attention fusion module (DAF), which can combine the restored and known regions in an adaptive manner. This multi-scale architecture is beneficial for the decoder to utilize discriminative representations learned from encoders into images. Extensive experiments clearly demonstrate that the proposed approach not only achieves more appealing results compared with state-of-the-art methods but also improves the performance of masked face recognition dramatically. 
\end{abstract}

\section{Introduction}

Face completion (a.k.a face inpainting or face hole-filling) aims at filling missing regions of a face image with plausible contents \cite{bertalmio2000image}. It is more difficult than general image inpainting because there are high-level identity information, pose variations, etc in face images. Face completion is a fundamental low-level vision task and can be applied to many downstream applications, such as photo editing and face verification \textcolor{black}{\cite{yu2019free,barnes2009patchmatch,wang2019laplacian}}. The target of face completion is to produce semantically meaningful content and reasonable structure information in missing areas.


There are many attempts for face completion, but they usually treat it as a general image inpainting problem. Traditional image inpainting methods \textcolor{black}{\cite{barnes2009patchmatch,efros2001image,xu2010image}} (e.g., PatchMatch) assume that the content to be filled comes from the background area. Therefore, they gradually synthesize plausible stationary contents by copying and pasting similar patches from known areas. The performances of these methods are satisfying when dealing with background inpainting tasks. But non-repetitive and complicated scenes, such as faces and objects, are the Waterloo of these traditional methods because of the limited ability to capture high-level semantics. Recently, deep convolutional neural networks (CNNs) have made great progress in many computer vision tasks \textcolor{black}{\cite{luo2021partial, jia2021inconsistency, ding2021unsupervised, luo2021fa, huang2019wavelet, pei2021all, he2020non}}. Thus, many deep learning-based methods have been proposed. Benefiting from the powerful ability of representation learning of CNNs, their performance has been significantly improved. These approaches adopt autoencoder or its variant architectures jointly trained with generative adversarial networks (GANs) to hallucinate semantically plausible contents in missing regions \cite{yu2019free,xie2019image,liu2018image}. But these methods still suffer from three problems\textcolor{black}{:}


Firstly, various kinds of masks are often presented in face images in the wild, especially in this tough period of COVID-19, which greatly increases the difficulty of image inpainting. Previous image inpainting approaches usually train an encoder and a decoder jointly with some commonly-used loss functions (e.g., reconstruction loss, style loss, etc). But encoders still struggle to learn powerful representations from images with various kinds of masks. As a result, these CNN-based approaches will produce unsatisfactory results with obvious artifacts. \textcolor{black}{A naive solution is to design a very deep network to obtain a large model capacity for learning powerful representations.} However, it will increase the computational cost heavily and may not help to learn accurate latent representations.


To cope with this limitation, we propose a self-supervised Siamese inference network with contrastive learning. \textcolor{black}{We assume that two identical images with different masks form a positive pair while a negative pair consists of two different images. Contrastive learning aims to maximize (minimize) the similarities of positive pairs (negative pairs) in a representation space. As explored in \cite{he2020momentum, hadsell2006dimensionality}, contrastive learning can be regarded as training an encoder to perform a dictionary $look$-$up$ task. An encoded ‘query’ should be matched with its corresponding ‘key’ (token) and different from others. The ‘keys’ (tokens) in the dictionary are usually sampled from images, patches, or other data types.} \textcolor{black}{In order to acquire a large and consistent dictionary, we design a queue dictionary and a momentum-updated key encoder. As demonstrated in MoCo \cite{he2020momentum}, the proposed self-supervised inference network can learn good features from input images. Thus, the robustness and the accuracy of the encoder can be improved.}


Secondly, previous methods consider image inpainting as a conditional image generation task. The roles of the encoder and decoder are recognizing high-level semantic information and synthesizing low-level textures \cite{yu2018generative}, respectively. These approaches, e.g., PConv \cite{liu2018image} and LBAM \cite{xie2019image}, focus more on missing areas and synthesize realistic alternative contents by a well-designed architecture or some commonly-used loss functions. However, there are either obvious color contrasts or artificial edge responses, especially in the boundaries of results produced by these methods since they ignore the structural consistency. \textcolor{black}{In fact, the development of biology has revealed that the human visual system is more sensitive to the topological distinction \cite{chen1982topological}}. Therefore, we focus not only on the structural continuity of restored images surrounding holes but also on generating texture-rich images. 

To properly suppress color discrepancy and artifacts in boundaries, we propose a novel dual attention fusion module (DAF) to synthesize pixel-wise smooth contents, which can be inserted into autoencoder architectures in a plug-and-play way. The core idea of the fusion module is to calculate the similarity between the synthesized content and the known region. Some methods are proposed to address this problem, such as DFNet \cite{hong2019dfnet} and Perez's method \cite{perez2003poisson}. However, these methods lack flexibility in handling different information types (e.g., different semantics), hindering learning more discriminative representations. Our proposed DAF is developed to adaptively recalibrate channel-wise features by taking interdependencies between channels into account and force CNNs to focus more on unknown regions. DAF will predict an adaptive spatial attention map to blend restored contents and original images naturally.

Finally, the verification performance heavily relies on the pixel level similarity and feature level similarity according to \cite{zhang2017demeshnet}, which means that the geometric information of the output results should be similar to the input. In practice, face appearance will be influenced by a number of factors such as meshes, wearing masks \cite{li2020learning,zhang2017demeshnet,cai2020semi} and so on. Masks can significantly destroy the facial shape and geometric information, greatly increasing the difficulty of generating visually appealing results. Therefore, it inevitably leads to a sharp decline in face verification performance. For example, healthcare workers must wear sanitary masks to avoid infection of diseases, and they will fail to pass through the face verification system. 

In this paper, we assume that the geometric information of the input face image should be kept intact. Inspired by recent advances in 3D face analysis \cite{alp2017densereg, alp2018densepose}, a dense correspondence field estimation is integrated into our network since it contains the complete geometric information of the input face. For simplicity, instead of using another network to predict the dense correspondence field separately, we make our decoder simultaneously predict the dense correspondence field and feature maps at multi-scales. Thus, we subtly employ a 3D supervision for our network provided by the dense correspondence field. \textcolor{black}{Under this 3D geometric supervision, our network can generate inpainting results with reasonable structure information.}

Qualitative and quantitative experiments are conducted on multiple datasets to evaluate our proposed method. The experimental results demonstrate that our proposed method not only outperforms state-of-the-art methods in generating high-quality inpainting results but also improves the performance of masked face recognition dramatically.

This paper is an extension of our previous conference publication \cite{ma2020free}. We extend it in three folds: 1) A dense correspondence field is proposed to be integrated into our network for utilizing 3D prior information of human faces. It can help our network to retain the facial shape and appearance information from the input. 2) We mainly concentrate on face image completion rather than other types of images. We add an extra face dataset, Flickr-Faces-HQ (FFHQ) \cite{karras2019style}, to demonstrate the effectiveness of our method. 3) We conduct an identity verification evaluation for face completion. It clearly shows the advantage of the proposed method compared with state-of-the-art methods.

To sum up, the main contributions of this paper are as follows:
\begin{itemize}
    \item We propose a Siamese inference network based on contrastive learning for face completion. It helps to improve the robustness and accuracy of representation learning for complex mask patterns.
    \item We propose a novel dual attention fusion module that can explore feature interdependencies in spatial and channel dimensions and blend features in missing regions and known regions naturally. Smooth contents with rich texture information can be naturally synthesized.
    \item To keep structural information of the input intact, the dense correspondence field that binds 2D and 3D surface spaces is estimated in our network, which can preserve the expression and pose of the input.
    \item Our proposed method achieves smooth inpainting results with rich texture and reasonable topological structural information on three standard datasets against state-of-the-art methods, and also greatly improves the performance of face verification. 
\end{itemize}

\section{Related Work}
\subsection{Image Inpainting}
Image inpainting aims to generate alternative contents when a given image is partially occluded or corrupt. Early traditional image inpainting methods are mainly diffusion-based \cite{bertalmio2000image} or patch-based \cite{barnes2009patchmatch}. They often use the information of the pixels (or image patches) around the occluded area to fill the missing regions. Bertalmio \textit{et al.} \cite{bertalmio2000image} proposed an algorithm to fill missing regions with information surrounding them automatically based on the principle that isophote lines arriving at the boundaries of the regions are completed inside. Barnes \textit{et a.} \cite{barnes2009patchmatch} presented a fast nearest neighbor searching algorithm named PatchMatch, to search and paste the most similar image patches from the known regions. These methods utilize low-level image features to guide the feature propagation from known image backgrounds or image datasets to corrupted regions. Criminisi \textit{et al.} \cite{criminisi2004region} proposed an efficient algorithm, which combined the advantages of 'texture synthesis' techniques and 'inpainting' techniques. Specifically, they designed a best-first method to find the most similar patches and used them to recover the corrupted regions gradually. These methods work well when holes are small and narrow, or there are plausible matching patches in uncorrupted regions. However, when suffering from complicated scenes, it is difficult for these approaches to produce semantically plausible solutions, due to a lack of semantic understanding of images.

Nowadays, deep learning techniques have made great contributions to computer vision communities. In order to accurately recover corrupted images, many methods adopt deep convolutional neural networks (CNNs) \textcolor{black}{\cite{wang2020multistage,ding2018perceptually}}, especially generative adversarial networks (GANs) \cite{goodfellow2014gan} in image inpainting. Pathak \textit{et al.} \cite{pathak2016contextencoder} formulates image inpainting as a conditional image generation problem. Then, they proposed a Context Encoder to recover corrupted regions according to surrounding pixels. Iizuka \textit{et al.} \cite{iizuka2017globalandlocal} utilized two discriminators to improve the quality of the generated images at different scales, facilitating both globally and locally consistent image completion. At the same time, some approaches designed a coarse-to-fine framework to solve the sub-problem of image inpainting in different stages \cite{yu2018contextualattention, Nazeri_2019_ICCV, ren2019structureflow}. Nazeri \textit{et al.} \cite{Nazeri_2019_ICCV} proposed to firstly recover the edge map of the corrupted image, then generate image textures in the second stage. Ren \textit{et al.} \cite{ren2019structureflow} proposed a method in which a structure reconstructor was employed to generate the missing structures of the inputs while a texture generator yielded image details. Zhang \textit{et al.} \cite{zeng2020high} proposed an iterative inpainting approach that contained a corresponding confidence map in results. They used this map as feedback and recovered holes by trusting high-confidence pixels.


\begin{figure*}[ht]
    \centering
    \includegraphics[scale=0.6]{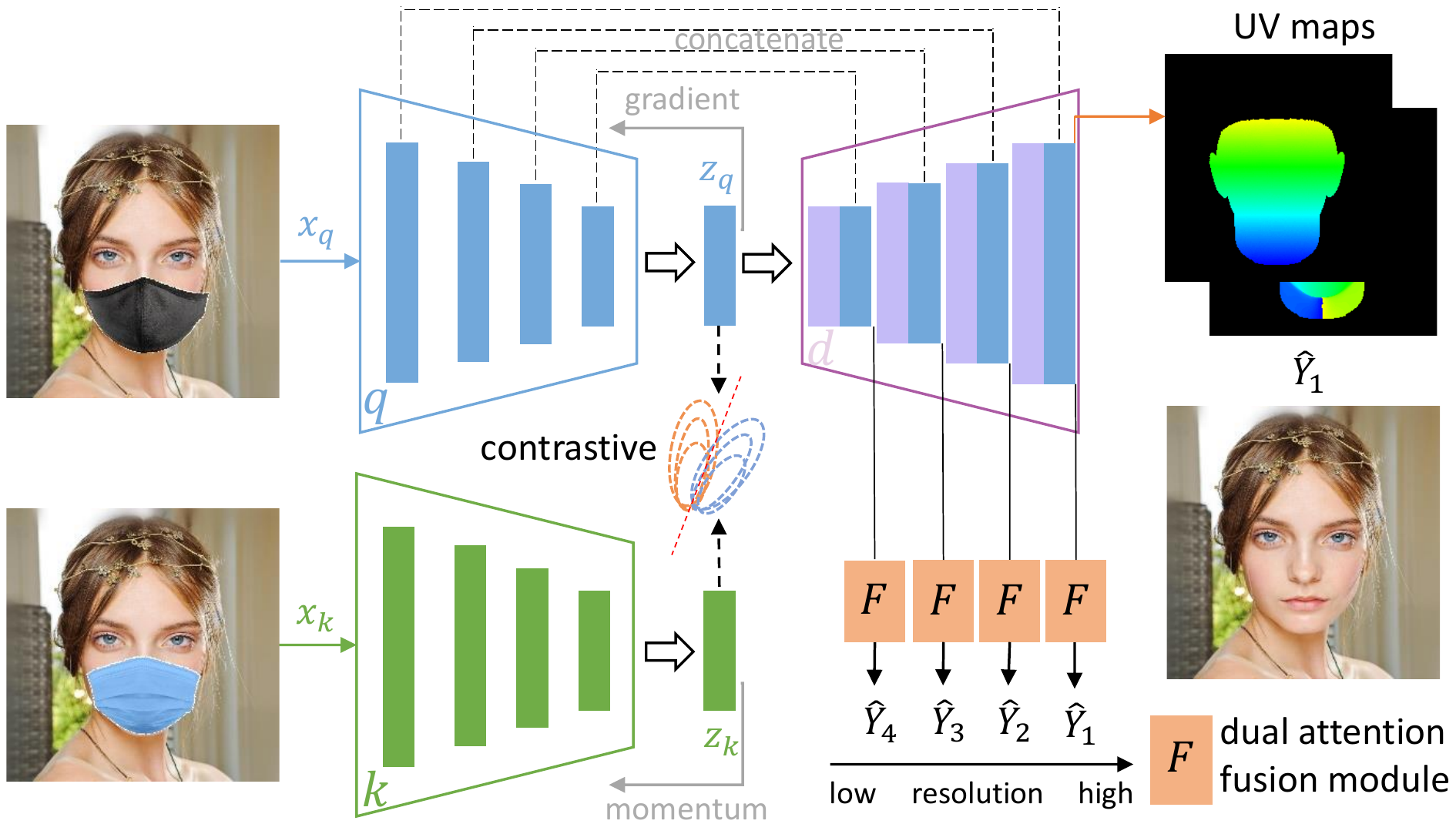}
    \caption{The network architecture of our method. The self-supervised Siamese inference network consists of encoders $E_q$ and $E_k$. This inference network encodes the new key representations on-the-fly by using the momentum-updated encoder $E_k$. We insert the dual attention fusion module into several decoder layers, forming a multi-scale decoder. We allow the decoder to estimate the dense correspondence field and the feature maps that are used for the DAF module at multi-scales simultaneously. The inference network is firstly trained with contrastive learning. Then the pre-trained encoder $E_q$ and the decoder are jointly trained with the fusion module.}
    \label{pipeline architecture}
\end{figure*}

As a branch of image inpainting, face completion is different from general image inpainting since its target mainly focuses on restoring the topological structure and texture of the face input. Zhang \textit{et al.} \cite{zhang2017demeshnet} argued that the performance of verification relied on both the pixel level similarity and the feature level similarity. Therefore, they proposed a feature-oriented blind face inpainting framework. Cai \textit{et al.} \cite{cai2019fcsr} proposed a method named FCSR-GAN to perform face completion and face super-resolution by multi-task learning where the generator was required to generate a high-resolution face image without occlusion from the occluded low-resolution face image. Zhou \textit{et al.} \cite{zhou2020learning} argued that previous works overlooked the serious impacts of inaccurate attention scores. Thus, they integrated the oracle supervision signal into the attention module to produce reasonable attention scores.

\subsection{Unsupervised Representation Learning}
Unsupervised learning has shown great potential to learn powerful representations of images recently \cite{he2020momentum, zhan2020self, chen2020SimCLR}. Compared with supervised learning, unsupervised learning utilizes unlabeled data to learn representations, which can go back to as far as the literature proposed by Becker and Hinton\cite{becker1992self}. Dosovitskiy \textit{et al.} \cite{dosovitskiy2014discriminative} proposed to discriminate between a set of surrogate classes generated by applying a number of transformations. Wu \textit{et al.} \cite{wu2018unsupervised} treated instance-level discrimination as a metric learning problem. Then, the discrete memory bank was utilized to store the features for each instance. Zhuang \textit{et al.} \cite{zhuang2019local} maximized a dynamic aggregation metric, which can move similar data instances together in the embedding space and separate dissimilar instances. He \textit{et al.} \cite{he2020momentum} proposed a dynamic dictionary consisting of a queue encoder and a moving-averaged encoder from a perspective on contrastive learning and they called this method MoCo. At the same time, Chen \textit{et al.} \cite{chen2020SimCLR} also presented a simple framework with contrastive learning for visual representations (SimCLR). Technically, they simplified recent contrastive learning-based algorithms and did not require specific structures and memory banks. Unsupervised learning strategies are also used in many computer vision tasks recently. Mustikovela \textit{et al.} \cite{mustikovela2020self-viewpoint} used self-supervised learning for viewpoint estimation by making use of generative consistency and symmetry constraint. Zhan \textit{et al.} \cite{zhan2020self-occlusion} utilized a mask completion network to predict occlusion ordering with a self-supervised learning strategy.

\subsection{Attention Mechanism}
Attention mechanism is a hot topic in computer vision and has been widely investigated in many works \textcolor{black}{\cite{wang2018non-local,chen2022multi, obeso2021visual, ding2018image}}. The wildly-used attention mechanism can be coarsely divided into two categories: spatial attention \cite{wang2018non-local} and channel attention \cite{hu2018senet} for image inpainting. Yu \textit{et al.} \cite{yu2018contextualattention} argued that convolutional neural networks lacked the ability to borrow or copy information from distant places, which led to blurry textures in generated images. Thus, they proposed a contextual attention module to calculate the spatial attention scores between pixels in the corrupted region and known region. Hong \textit{et al.} \cite{hong2019dfnet} proposed a fusion block to generate an adaptive spatial attention map $\alpha$ to combine features in the corrupted region and known region. In this paper, we investigate both spatial attention and channel attention mechanism to further improve the performance of face completion.

\subsection{3D Face Analysis}
Nowadays, the famous 3DMM \cite{blanz1999morphable} is widely used to express facial shape and appearance information for face related tasks, such as facial attribute editing, face hallucination, etc \cite{li2019disentangled, tu2019joint}. Roth \textit{et al.} \cite{roth2015unconstrained} proposed a photometric stereo-based method for unconstrained 3D face reconstruction, which benefited from a combination of landmark constraints and photometric stereo-based normals. Yin \textit{et al.} \cite{dosovitskiy2014discriminative} proposed a generative adversarial network combined with 3DMM, termed as FF-GAN, to provide shape and appearance priors without requiring large training data. 2DASL \cite{tu20203d} utilized 2D face images with noisy landmark information in the wild to assist 3D face model learning. It has become a popular method to establish the dense correspondence field between the 2D and 3D space. Güler \textit{et al.} \cite{alp2017densereg, alp2018densepose,cao2018learning} proposed a UV correspondence field to build pixel-wise correspondence between RGB color space and 3D surface space. These works show that the UV correspondence field can retain geometric information of the human face.

\section{Methodology}
In this section, we first present our self-supervised Siamese inference network. Subsequently, the details of the dual attention fusion (DAF) module, the dense correspondence estimation, and learning objectives in our method are provided. The overall framework of our face completion method is shown in Fig. \ref{pipeline architecture}.

\subsection{Self-Supervised Siamese inference network}
Our proposed self-supervised Siamese inference network consists of two identical encoders but not sharing parameters \cite{he2020momentum,hadsell2006dimensionality,wang2015unsupervised}, noted as $E_q$ and $E_k$, respectively. The proposed inference network is trained with contrastive learning, which can be viewed as training an encoder to perform a dictionary look-up task: a 'query' encoded by $E_q$ should be similar with its corresponding 'key' (i.e., positive key) represented by another encoder $E_k$ and dissimilar to others (i.e., negative keys). Two images with different masks are required for the proposed inference network, named as $x_q$ and $x_k$, respectively. Thus, we can obtain a query representation $z_q=E_q(x_q)$ and a key representation $z_k=E_k(x_k)$, respectively. Following many previous self-supervised works \cite{zhuang2019local,bachman2019learning}, the contrastive loss is utilized as the self-supervised objective function for training the proposed inference network and can be written as:
\begin{equation}
    \mathcal{L}=-log\frac{exp(z_q.z_k^+/\tau)}{\sum_{i=0}^{K}exp(z_q.z_{k_i}/\tau)},
\end{equation}
where $\tau$ is the temperature hyper-parameter, and the loss function will degrade into the original $softmax$ when $\tau$ is equal to 1. The output will be less sparse with $\tau$ increasing \cite{chen2020dynamic}. The $\tau$ is set as $0.07$ for the efficient training process in this work. Specially, this loss, also known as InfoNCE loss\cite{he2020momentum,hadsell2006dimensionality}, tries to classify $z_q$ as $z_k^+$. Here, $z_q$ and $z_k^+$ are encoded from a positive pair. $K$ means the number of negative samples.

High-dimensional continuous images can be projected into a discrete dictionary by contrastive learning. There are three general mechanisms for implementing contrastive learning (i.e., end-to-end training \cite{hadsell2006dimensionality}, memory bank \cite{wu2018unsupervised} and momentum updating \cite{he2020momentum}), whose main differences are how to maintain keys and how to update the key encoder. Considering GPU memory size and powerful feature learning, we follow MoCo \cite{he2020momentum} to design a consistent dictionary implemented by $queue$. Thus, the key representations of the current batch data are enqueued into the dictionary while the oldest representations are 
dequeued progressively. The length of the queue is under control, which enables the dictionary to contain a large number of negative image pairs. Such a dictionary with large-scale negative pairs will facilitate representation learning. We set the length of the queue as $65536$ in this work.

It is worth noting that the encoder $E_k$ is updated by a momentum-updated strategy instead of direct back-propagation. The main reason is that it's difficult to propagate the gradients to all keys in the queue. The updating process of $E_k$ can be formulated as follows:
\begin{equation}
    \theta_k \leftarrow m\theta_k+(1-m)\theta_q,
\end{equation}
where $\theta_q$ and $\theta_k$ denotes as the parameters of $E_q$ and $E_k$, respectively. $\theta_q$ is updated by back-propagation. $m\in[0,1)$ is the momentum coefficient hyper-parameter and set as $0.9$ in this paper. \textcolor{black}{The momentum-update mechanism makes the encoder $E_k$ update smoothly relative to $E_q$, resulting in a more consistent discrete dictionary.}

\subsection{Dual Attention Fusion Module}
\label{dual attention fusion module}
\begin{figure}[ht]
    \centering
    \includegraphics[scale=0.55]{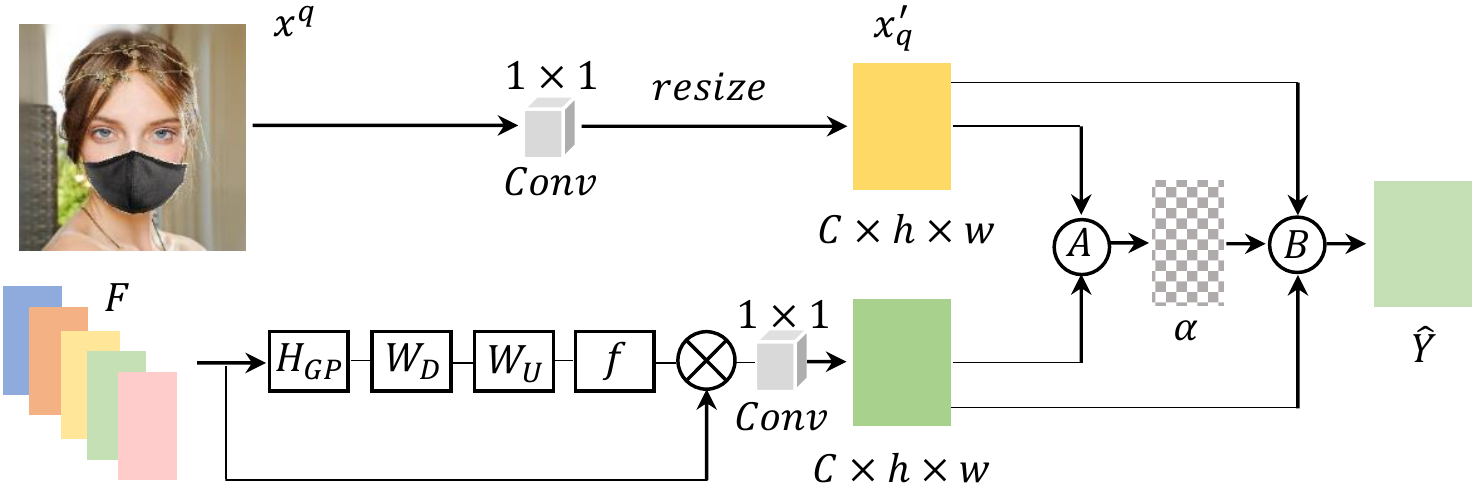}
    \caption{The architecture of the dual attention fusion module. It firstly predicts an adaptive spatial attention map $\alpha$ with the learnable transformation function $\mathcal{A}$. Then we can obtain final natural face completion results with rich texture by the fusion function $\mathcal{B}$.}
    \label{fusion}
\end{figure}

We now give more details about our proposed dual attention fusion module (see Fig. \ref{fusion}), which contains a channel attention mechanism and a spatial attention mechanism. This fusion module is embedded into the last several layers of the decoder and outputs face completion results with multi-scale resolutions \textcolor{black}{\cite{karras2017progressive}}. Thus, constraints can be imposed on multi-scale outputs for high-quality results.

Previous CNN-based image inpainting approaches treat channel-wise features equally, thus hindering the ability of the representation learning of the network. Meanwhile, high-level and interrelated channel features can be considered as specific class responses. For more discriminative representations, we first build a channel attention module in our proposed fusion module. 

As shown in Fig. \ref{fusion}, let a feature map $F=[f_1,\cdots,f_c,\cdots,f_C]$ be one of the inputs of the fusion module, whose channel index is $c$ and size is $h \times w$. The channel descriptor can be acquired from the channel-wise global spatial information by global averaging pooling. Then we can obtain the channel-wise statistics $z_c \in \mathbb{R}^c$ by shrinking $F$:
\begin{equation}
    z_c=H_{GP}(\textcolor{black}{f_c})=\frac{1}{h \times w}\sum_{i=1}^{h}\sum_{j=1}^{w}f_c(i,j),
\end{equation}
\textcolor{black}{where $z_c$ is the $c$-th element of $z$. $f_c(i,j)$ is the value at position $(i,j)$ of $c$-th feature $f_c$. $H_{GP}$ means the global pooling function.}

In order to fully explore the channel-wise dependencies of the aggregated information, we introduce a gating mechanism. As illustrated in \cite{hu2018senet,zhang2018image}, the sigmoid function can be used as a gating function:
\begin{equation}
    \omega=\textcolor{black}{\sigma}(W_U\delta(W_Dz)),
\end{equation}
where $\textcolor{black}{\sigma(\cdot)}$ and $\delta(\cdot)$ are the sigmoid gating and ReLU functions, respectively. $W_D$ and $W_U$ are the weight sets of $Conv$ layers who set channel number as $C/r$ and $C$, respectively. Finally, the channel statistics $\omega$ are acquired and used to rescale the input $f_c$:
\begin{equation}
    \hat{f_c} = w_c\cdot f_c,
\end{equation}
where $w_c$ and $f_c$ are the scaling factor and feature map of the $c$-th channel, respectively.

The long-range contextual information is essential for discriminant feature representations. We propose a spatial attention module that forms the final part of the proposed fusion module. Given an input image with a mask $x_q$, we first get $x{_q}^{'}$ that matches the size of the re-scaled feature map $\hat{F} \in \mathbb{R}^{c \times h \times w}$,
\begin{equation}
    x{_q}^{'}=(W_Cx_q)\downarrow,
\end{equation}
where $W_C$ and $\downarrow$ are the weight set of a $1 \times 1$ convolutional layer and downsample module, respectively. 

Then the adaptive spatial attention map $\alpha \in \mathbb{R}^{C \times h \times w}$ is given by,
\begin{equation}
    \alpha=\textcolor{black}{\sigma}(\mathcal{A}(\textcolor{black}{W_K}\hat{F},x{_q}^{'})),
\end{equation}
where $\textcolor{black}{W_K}$ is the weight set of a $1 \times 1$ convolutional layer. It sets channel number of $\hat{F}$ to be same with $x{_q}^{'}$. $\mathcal{A}$ is a learnable transformation function implemented by three $3 \times 3$ convolutional layers. $\textcolor{black}{W_K}\hat{F}$ and $x{_q}^{'}$ are first concatenated and then fed into the convolutional layers. $f(\cdot)$ is the sigmoid function that can make $\alpha$ an attention map to some extent.

The final inpainting result $\hat{Y}$ is obtained by, 
\begin{equation}
    \hat{Y}=\mathcal{B}(\alpha,\textcolor{black}{W_K}\hat{F},x{_q}^{'})=\alpha \odot \textcolor{black}{W_K} \hat{F}+(1-\alpha)\odot{x}{_q}^{'},
\end{equation}
\textcolor{black}{where $\odot$ and $\mathcal{B}$ denote the Hadamard product and fusion function, respectively. The adaptive spatial attention map $\alpha$ can adjust the balance between the ground truth image and the restored image to obtain a smoother transition. We can eliminate obvious color contrasts and artifacts especially in boundary areas, and get natural face completion results with richer textures.}

\begin{figure}[ht]
    \centering
    \includegraphics[scale=0.6]{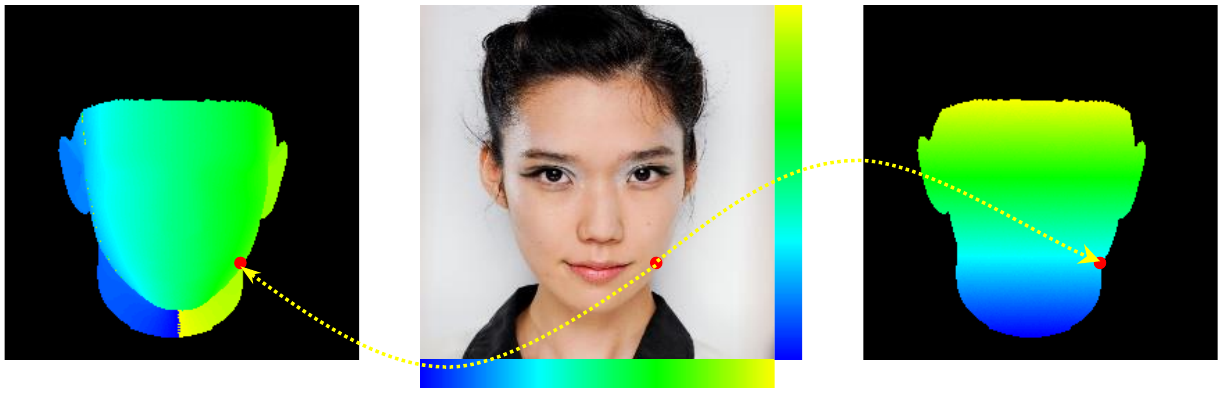}
    \caption{Visualization examples of the dense correspondence. The face image is shown in the middle. The corresponding U map and V map are shown in the left and right, respectively.}
    \label{UV map}
\end{figure}

\subsection{Dense Correspondence Field Estimation}
\label{dense correspondence field estimation}
Masks can dramatically destroy the facial shape and structure information, such as viewing angles, facial expressions, and so on, making it quite tough to achieve visually appealing results. To keep the geometric information of the human face intact during the face completion process, we introduce a dense correspondence field that binds the 2D and 3D surface spaces into our network.

The structure and texture information of the face image can be disentangled by the dense correspondence field according to \cite{alp2017densereg,alp2018densepose}. The geometrical information is stored in the correspondence field while the texture map can represent the surface of a 3D face to some extent. In this paper, we mainly concentrate on inferring the dense correspondence field by our network. Technically, given an input image $x \in \mathbb{R}^{c \times h \times w}$, the dense correspondence field $C = (u;v)$ consists of maps in the UV space ($u, v \in \mathbb{R}^{h \times w}$). The visual illustration is shown in Fig. \ref{UV map} in which the minimum is rendered as blue and the maximum is rendered as yellow.

We allow our decoder to predict the dense correspondence fields and feature maps at multi-scales simultaneously, where the feature maps are fed into the proposed dual attention fusion module (please see Sec. \ref{dual attention fusion module}). Thanks to the multi-scale network architecture, our decoder can obtain context information better and maintain geometrical information. In order to supervise $C$ during training, we minimize the pixel-wise error between the estimated result and the ground truth $C$. It can be written mathematically as,
\begin{equation}
    \mathcal{L}_{UV} = ||C^{'} - C||_2,
\end{equation}
where $C^{'}$ means the predicted dense correspondence field result of an input image. We employ BFM \cite{paysan20093d}, a 3D shape estimation approach, to obtain the ground truth dense correspondence field $C$ similar with \cite{li2019disentangled,cao2018learning}. We then obtain coordinates of vertices by performing the model fitting method \cite{zhu2016face}. Finally, those vertices are mapped to the UV space by the cylindrical unwrapping according to \cite{booth2014optimal}.

\subsection{Loss Functions}
Following \textcolor{black}{\cite{ding2018perceptually,zeng2021feature,kim2019recurrent}}, for synthesizing richer texture details and correct semantics, the element-wise reconstruction loss, the perceptual loss \cite{johnson2016perceptual}, the style loss and the adversarial loss are used in our proposed method. Moreover, we also employ the identity preserving loss function to ensure that the identity information of the generated images remains unchanged.

\textbf{Reconstruction Loss.} It is calculated as $\mathcal{L}_1$-norm between the inpainting result $\hat{Y}$ and the target image $Y$,
\begin{equation}
    \mathcal{L}_{rec}=||Y-\hat{Y}||_1.
\end{equation}


\textbf{Style Loss.} For getting richer textures, we also adopt the style loss defined on the feature maps produced by the pre-trained VGG-16. Following \cite{xie2019image,liu2018image}, the style loss can be calculated as the $\mathcal{L}_1$-norm between the Gram matrices of the feature maps,
\begin{equation}
    \mathcal{L}_{style}=\frac{1}{N}\sum^N_{i=1}\frac{1}{C_i \cdot C_i}||\Phi^i(Y)(\Phi^i(Y))^T-\Phi^i(\hat{Y})(\Phi^i(\hat{Y}))^T||_1,
\end{equation}
where $C_i$ denotes the channel number of the feature map at $i$-th layer in the pre-trained VGG-16.

\textbf{Identity Preserving Loss.} To ensure the generated face images belong to the same identity as the target face images, we adopt LightCNN \cite{wu2018light} to extract the features, then use the mean square error to constrain the embedding spaces,
\begin{equation}
    \mathcal{L}_{ip}=||\Psi(Y)-\Psi(\hat{Y})||_2,
\end{equation}
where $\Psi$ means the pre-trained LightCNN network \cite{wu2018light}.


\textbf{Model Objective.} The above loss functions can be grouped into two categories: $Structure$ $Loss$ and $Texture$ $Loss$, respectively. The $Structure$ $Loss$ is given by, 
\begin{equation}
    \mathcal{L}_{struct}^k=\lambda_{rec}\mathcal{L}_{rec}^k + \lambda_{uv}\mathcal{L}_{uv}^k,
\end{equation}
where $\lambda_{rec}$ and $\lambda_{uv}$ mean the weight factors and are set as $6$ and $0.1$ empirically. $\mathcal{L}_{struct}^k$ is calculated as the sum of $\mathcal{L}_{rec}$ and $\mathcal{L}_{uv}$ at the $k$-th layer of the decoder. Here, $\mathcal{L}_{uv}$ means the UV loss function (please see Sec. \ref{dense correspondence field estimation}). 

The $Texture$ $Loss$ is given by,

\begin{equation}
    \mathcal{L}_{text}^{k}=\lambda_{style}\mathcal{L}_{style}^{k}+\lambda_{ip}\mathcal{L}_{ip}^{k},
\end{equation}
where $\lambda_{style}$ and $\lambda_{ip}$ are trade-off factors and are set as $240$ and $0.1$ empirically in this work.

Finally, the total model objective can be formulated as,
\begin{equation}
    \mathcal{L}_{total}=\frac{1}{|P|}\sum_{k \in P}\mathcal{L}_{struct}^k+\frac{1}{|Q|}\sum_{k \in Q}\mathcal{L}_{text}^k,
\end{equation}
where both $P$ and $Q$ are the selected decoder layer sets that imposed constraints. We select $P$ as $\{1,2,3,4,5,6\}$ and $Q$ as $\{1,2,3\}$ respectively for better inpainting results. Note that $1$ represents the outermost layer.

\section{Experiments}
To demonstrate the superiority of our approach against state-of-the-art methods, both quantitative and qualitative experiments for face completion and face verification experiments are conducted. In this section, we will introduce the details of our experimental settings and the experimental results one by one.

\begin{figure*}[ht]
    \centering
    \includegraphics[scale=0.45]{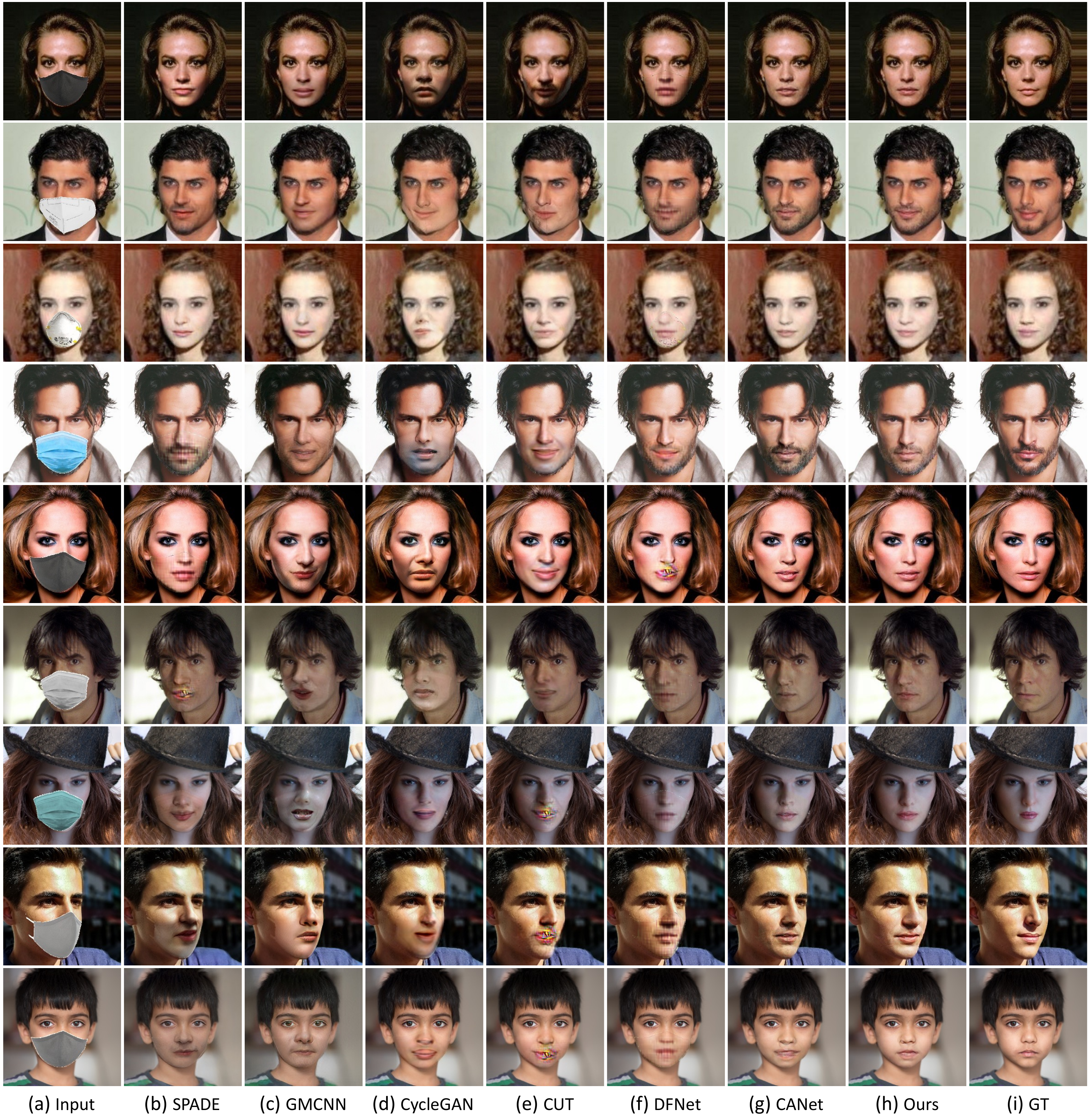}
    \caption{Qualitative results compared with state-of-the-arts on three datasets. From left to right, (a) are the input images with various kind of masks. (b), (c), (d), (e), (f), (g) and (h) are the results generated by SPADE \cite{park2019semantic}, GMCNN \cite{wang2018image}, CycleGAN \cite{zhu2017unpaired}, CUT \cite{park2020contrastive}, DFNet \cite{hong2019dfnet}, CANet \cite{ma2020free} and ours method respectively. (i) is the ground truth.}
    \label{comaparison experments}
\end{figure*}

\begin{table*}[]
\begin{center}
\begin{tabular}{c|ccc|ccc|ccc}
\hline
Dataset  & \multicolumn{3}{c|}{CelebA} & \multicolumn{3}{c|}{CelebA-HQ} & \multicolumn{3}{c}{FFHQ} \\ \hline
Metric   & PSNR\ddag    & SSIM\ddag    & FID\dag     & PSNR\ddag     & SSIM\ddag      & FID\dag     & PSNR\ddag   & SSIM\ddag    & FID\dag    \\ \hline\hline
SPADE \cite{park2019semantic}   & 30.92   & 0.9640  & 1.8216  & 27.40    & 0.9321    & 30.07   & 26.27  & 0.9170  & 24.45  \\
GMCNN \cite{wang2018image}   & 29.91   & 0.9563  & 2.6205  & 26.10    & 0.9107    & 13.07   & 25.30  & 0.8963  & 8.72   \\
CycleGAN \cite{zhu2017unpaired} & 24.47   & 0.9063  & 4.9871  & 21.94    & 0.8446    & 13.75   & 21.13  & 0.8239  & 11.26  \\
CUT \cite{park2020contrastive}     & 24.55   & 0.9115  & 5.4059  & 22.65    & 0.8690    & 15.58   & 21.68  & 0.8429  & 12.75  \\
DFNet \cite{hong2019dfnet}   & 32.18   & 0.9706  & 4.2948  & 28.90      & 0.9494       & 8.40   & 29.33  & 0.9453  & 11.94  \\
\textcolor{black}{CANet \cite{ma2020free}}   &\textcolor{black}{32.49}   &\textcolor{black}{0.9731}   &\textcolor{black}{0.9778}   &\textcolor{black}{29.89}   &\textcolor{black}{0.9545}   &\textcolor{black}{4.23}   &\textcolor{black}{29.70}   &\textcolor{black}{0.9501}   &\textcolor{black}{2.12}   \\
Ours     & \textbf{33.26}   & \textbf{0.9769}  & \textbf{0.7981}  & \textbf{30.67}    & \textbf{0.9607}    & \textbf{3.53}    & \textbf{30.42}  & \textbf{0.9580}  & \textbf{1.75}   \\ \hline
\end{tabular}
\end{center}
\caption{Quantitative comparison on the testing sets of CelebA, CelebA-HQ and FFHQ. \dag Lower is better. \ddag Higher is better.}
\label{quantitative results}
\end{table*}

\subsection{Datasets and Protocols}
\label{Datasets and Protocols}
\textbf{CelebA.} The CelebFaces Attributes dataset \cite{liu2015faceattributes} is widely used for face hallucination, image-to-image translation, etc. It's a large-scale face attributes dataset containing more than 200k celebrity images, which includes face images with large occlusion and pose variations. We randomly select 10,000 images for testing and the rest for training.

\begin{figure*}[ht]
    \centering
    \includegraphics[scale=0.5]{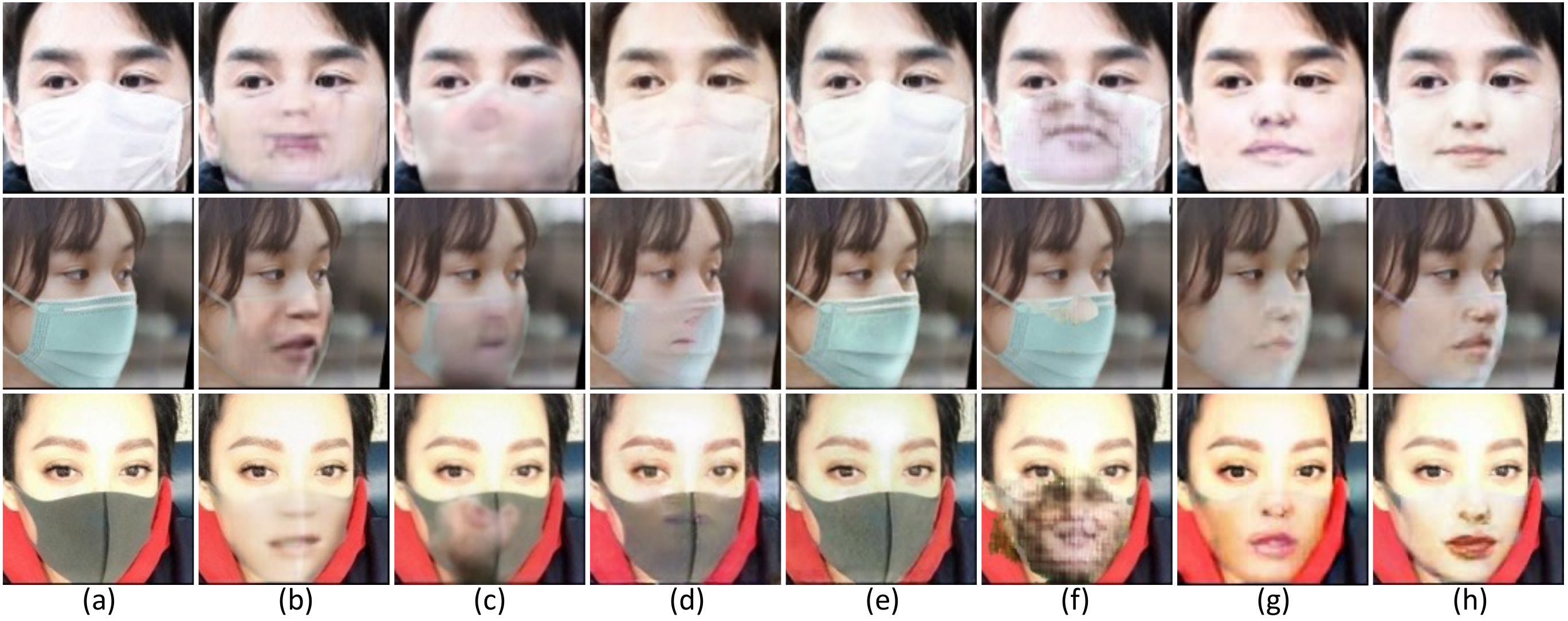}
    \caption{Face completion results in the wild. (a) is the inputs. (g) is the results generated by our method. (b) - (g) are results produced by SPADE, GMCNN, CycleGAN, CUT, DFNet and CANet, respectively.}
    \label{face completion results in the wild}
\end{figure*}

\begin{table*}[]
\centering
\begin{tabular}{cccccccc}
\hline
Methods  & SPADE  & GMCNN  & CycleGAN & CUT    & DFNet  & CANet & Ours   \\ \hline
FID \dag     & 113.52 & 150.98 & 167.73   & 179.78 & 173.41 & 103.34     & \textbf{98.39}  \\
LPIPS \dag   & 0.0827 & 0.1065 & 0.1116   & 0.1303 & 0.1208 & 0.0812     & \textbf{0.0709} \\
F1-Score \ddag & 0.026  & 0.0139 & 0.0022   & 0.0034 & 0.0022 & 0.0219     & \textbf{0.0493} \\
Realism \ddag  & 0.7613 & 0.7443 & 0.7089   & 0.6819 & 0.6759 & 0.7723     & \textbf{0.7883} \\ \hline
\end{tabular}
\caption{Quantitative comparison on the real world face dataset (RMFD). \dag Lower is better. \ddag Higher is better.}
\label{quantitative resutls in the wild}
\end{table*}

\begin{table*}[]
\centering
\begin{tabular}{cccccccc}
\hline
Methods & SPADE  & GMCNN  & CycleGAN & CUT    & DFNet  & CANet  & Ours            \\ \hline
PSNR \ddag   & 22.6   & 22.57  & 20.11    & 20.2   & 22.73  & 23.3   & \textbf{23.69}  \\
SSIM \ddag   & 0.8775 & 0.8785 & 0.8297   & 0.8292 & 0.8805 & 0.8975 & \textbf{0.9007} \\
FID \dag    & 40.59  & 44.31  & 64.09    & 62.12  & 33.04  & 31.6   & \textbf{29.14}  \\ \hline
\end{tabular}
\caption{Quantitative comparison on the L2SFO dataset. \dag Lower is better. \ddag Higher is better.}
\label{quantitative resutls on L2SFO}
\end{table*}

\begin{table*}[]
\centering
\begin{tabular}{cccccccccc}
\hline
Model                     & Metric    & Masked & SPADE & GMCNN & CycleGAN & CUT   & DFNet & CANet & Ours           \\ \hline
\multirow{3}{*}{ArcFace}  & AUC       & 97.46  & 97.60 & 97.68 & 97.56    & 97.57 & 97.66 & 97.77 & \textbf{98.03} \\
                          & FPR=1\%   & 80.07  & 80.98 & 81.60 & 81.18    & 81.14 & 81.61 & 82.23 & \textbf{84.38} \\
                          & FPR=0.1\% & 60.53  & 61.73 & 62.49 & 62.41    & 61.93 & 62.86 & 63.33 & \textbf{67.21} \\ \hline
\multirow{3}{*}{LightCNN} & AUC       & 99.13  & 99.16 & 99.15 & 99.14    & 99.14 & 99.14 & 99.14 & \textbf{99.20} \\
                          & FPR=1\%   & 93.50  & 93.90 & 93.70 & 93.68    & 93.56 & 93.68 & 93.88 & \textbf{94.62} \\
                          & FPR=0.1\% & 83.99  & 84.97 & 84.50 & 84.47    & 84.22 & 84.61 & 85.31 & \textbf{87.20} \\ \hline
\multirow{3}{*}{FaceNet}  & AUC       & 99.15  & 99.18 & 99.15 & 99.17    & 99.16 & 99.18 & 99.22 & \textbf{99.27} \\
                          & FPR=1\%   & 91.76  & 92.11 & 91.67 & 91.73    & 91.82 & 91.86 & 92.40 & \textbf{92.94} \\
                          & FPR=0.1\% & 78.47  & 79.15 & 78.43 & 78.72    & 78.59 & 79.14 & 79.72 & \textbf{80.88} \\ \hline
\end{tabular}
\caption{Face verification results on IJB-C. 'Masked' means face verification experiments are conducted between the masked probe set and the unchanged gallery set directly.}
\label{face-verification-on-IJB-C}
\end{table*}

\begin{table*}[]
\centering
\begin{tabular}{cccccccccc}
\hline
Model                     & Metric    & Masked & SPADE & GMCNN & CycleGAN & CUT   & DFNet & CANet & Ours  \\ \hline
\multirow{3}{*}{ArcFace \cite{deng2019arcface}}  & AUC       & 97.51  & 98.02 & 97.78 & 97.65    & 97.88 & 97.50 & 98.09 & \textbf{98.38} \\
                          & FPR=1\%   & 77.71  & 81.52 & 79.93 & 78.28    & 79.86 & 78.05 & 82.12 & \textbf{83.10} \\
                          & FPR=0.1\% & 44.85  & 45.56 & 43.33 & 44.38    & 46.43 & 45.79 & 50.07 & \textbf{56.84} \\ \hline
\multirow{3}{*}{LightCNN \cite{wu2018light}} & AUC       & 99.20  & 99.29 & 99.30 & 99.24    & 99.30 & 99.20 & 99.34 & \textbf{99.49} \\
                          & FPR=1\%   & 91.04  & 92.56 & 92.63 & 91.95    & 92.36 & 76.87 & 93.40 & \textbf{94.41} \\
                          & FPR=0.1\% & 77.13  & 74.41 & 77.78 & 76.06    & 80.34 & 64.85 & 80.34 & \textbf{82.73} \\ \hline
\multirow{3}{*}{FaceNet \cite{schroff2015facenet}}  & AUC       & 98.98  & 99.08 & 99.03 & 99.02    & 99.03 & 98.98 & 99.10 & \textbf{99.30} \\
                          & FPR=1\%   & 85.96  & 87.51 & 87.14 & 86.40    & 86.30 & 86.06 & 87.97 & \textbf{90.17} \\
                          & FPR=0.1\% & 55.56  & 57.07 & 53.10 & 54.31    & 56.06 & 55.72 & 56.03 & \textbf{57.85} \\ \hline
\end{tabular}
\caption{Face verification results on LFW. 'Masked' means face verification experiments are conducted between the masked probe set and the unchanged gallery set directly.}
\label{face verification}
\end{table*}

\textbf{CelebA-HQ.} It's a high-resolution face images dataset established by Karras \textit{et al.} \cite{karras2017progressive}, which contains 30,000 high-quality face images. We divide the dataset into two subsets: the training set of 28,000 images and the testing set of 2,000 images.

\textbf{FFHQ.} The Flickr-Faces-HQ dataset \cite{karras2019style} is a high-quality dataset containing 70,000 face images at $1024 \times 1024$ resolution. It also covers age, ethnicity, and image background variations. We randomly choose 6,000 images for testing and the rest for training.

\textbf{Multi-PIE.} It contains more than 750,000 images that cover 15 viewpoints, 19 illumination conditions and a number of facial expressions of 337 identities \cite{gross2010multi}. We follow Huang \textit{et al.} \cite{huang2017beyond} to split the dataset. In our experiments, we only utilize the training set to train our network and the compared methods for face recognition.

\textbf{LFW.} The Labeled Faces in the Wild \cite{huang2008labeled} is a benchmark database commonly used for face recognition, which contains 13,233 images of 5,749 people captured in unconstrained environments. LFW provides a standard protocol for face verification that contains 6,000 face image pairs (including 3,000 positive pairs and 3,000 negative pairs, respectively). We use these standard face image pairs to evaluate face verification performance via face completion. Specially, face images in the gallery set remain the same while the counterparts in the probe set are occluded by masks. We firstly recover the occluded face images by our proposed method and the state-of-the-arts. Then, we compare the verification performance. It's worth noting that we only use LFW for testing.

\textcolor{black}{\textbf{L2SFO.} It is a large-scale synthesized face-with-occlusion dataset built by Yuan \textit{et al.} \cite{yuan2019face}. We call it L2SFO in which face images are occluded by six common objects including masks, eyeglasses, sunglasses, cups, scarves, and hands. All the occlusions are located on face images according to segmentation information to augment the reality of this dataset. It contains 991 different identities and more than 73,000 images. We randomly select 891 identities as the training set (about 66,000 images) and the rest as the testing set (about 7,000 images).}

\textcolor{black}{\textbf{IJB-C.} IARPA Janus Benchmark C is a dataset consisting of video still-frames and photos and used for face recognition benchmark \cite{Whitelam2017IARPAJB}. It contains 117,500 frames from 11,799 videos and 3,531 subjects with 31,300 still images. We use the 1:1 protocol for face verification, whose probe and gallery templates are combined using some images and video frames for each subject. Same as the processing procedure of LFW, images in the probe set are occluded and images in the gallery set remain unchanged. We firstly generate clean face images from occluded face images by using our method and other compared methods and then compare the face verification performance. IJB-C is also only used for testing.}

\subsection{Implementation Details} 
In our experiments, face images are normalized to $256 \times 256$ and $128 \times 128$ for high-resolution face completion and face verification, respectively. Following Wu \textit{et al.} \cite{wu2018light}, the landmarks in the centers of the eyes and mouth are used for normalizing face images. The occluded face images are generated by MaskTheFace proposed by Anwar and Raychowdhury \cite{anwar2020masked}. We randomly select mask types to occlude face images during training. Some occluded face images are shown in Fig. \ref{comaparison experments} and Fig. \ref{multi}. For different experimental settings, different datasets are utilized to train our network. For face completion, we train our network on the training sets of CelebA, CelebA-HQ, FFHQ and L2SFO, then testing on their testing sets. As for face verification, we train our network on the training sets of CelebA and Multi-PIE and test on LFW and IJB-C.

Our proposed method can be broken down into two stages. In the first stage, the inference network is trained through contrastive learning until convergence. And in the next stage, the pre-trained encoder and the decoder are jointly trained with the fusion module. We use the SGD optimizer with the learning rate as 0.015 for training the Siamese inference network, and use the Adam optimizer with the learning rate as $10^{-4}$ for jointly training the encoder and decoder. All the results are reported directly without any additional post-processing. Our proposed method is implemented by the Pytorch framework and trained on four NVIDIA TITAN Xp GPUs (12GB).

\subsection{Face Completion Quantitative Results}
\label{face completion quantitative results}
Peak signal-to-noise ratio (PSNR), structural similarity index (SSIM), and Fr{\'e}chet  Inception Distance (FID) are used as evaluation metrics. PSNR and SSIM measure the similarity between the inpainting result and the target image. As for FID, it measures the Wasserstein-2 distance between real and inpainting images through the pre-trained Inception-V3. We select 'cloth \#333333', 'KN95', 'N95', 'surgical blue', 'cloth \#515151', 'surgical', 'surgical green', 'cloth \#dadad9' and 'cloth \#929292' masks to occlude the testing images for experiments. These mask images are shown in Fig. \ref{comaparison experments} from top to bottom.

We conduct quantitative experiments on the testing sets of CelebA, CelebA-HQ and FFHQ occluded by the nine kinds of masks, and report the averaged results. Table \ref{quantitative results} shows the performance of our proposed method against other state-of-the-art methods, which consists of two image inpainting methods, GMCNN \cite{wang2018image} and DFNet \cite{hong2019dfnet}, and three image-to-image translation methods: Spade \cite{park2019semantic}, CycleGAN \cite{zhu2017unpaired} and CUT \cite{park2020contrastive}. \textcolor{black}{In Table \ref{quantitative results}, we also conduct the experiments to show the improvement of performance compared to our prior conference work \cite{ma2020free}. For simplicity, we call it CANet, which can be regarded as a simplified version of our proposed method in this paper without \textit{Dense Correspondence Field Estimation} and the identity preserving loss.} We retrain all the compared methods on the training sets of CelebA, CelebA-HQ and FFHQ for the sake of fairness. As shown in Table \ref{quantitative results}, the proposed method and CANet achieve the best and the second-best quantitative results in three metrics on all the testing sets. The results suggest that the proposed method can generate very realistic face images while the compared methods may not work well encountered various kinds of masks. The main reasons for the relatively low performance of the compared methods (excluding CANet) are that 1): face images with various kinds of masks dramatically increase the difficulty of image inpainting, hindering the ability of the representation learning of the encoder; 2): exiting methods take generating realistic images into account but ignore the structural consistency of the generated image. \textcolor{black}{The reason why the performance of our method is higher than CANet may be that \textit{Dense Correspondence Field Estimation} keeps the geometric information of the human face intact during the face completion process.}

\subsection{Face Completion Qualitative Results}
We compare our proposed method with state-of-the-art methods in terms of visual and semantic coherence. We conduct qualitative experiments on the testing sets of three datasets with various kinds of masks. As shown in Fig. \ref{comaparison experments}, we mask the testing images with the nine kinds of masks as described in the last section. 

Among all these compared methods, there are severe artifacts in results produced by SPADE, CUT, and DFNet. Thus, the qualities of generated images are far from the requirements. The reason is that various kinds of masks hinder their networks to capture powerful representations. There are no obvious artifacts in face images produced by CycleGAN. But it fails to maintain the geometric information of face images and produce obvious color contrasts. The reason is that CycleGAN endeavors to translate the input to its correspondence non-mask face image and ignores the structural consistency. As for GMCNN, it produces relatively appealing results, but there are significant differences in color at the edges. \textcolor{black}{CANet produces better results in which the facial geometric information is maintained but there are still artifacts, especially in the corners of the mouth.} Compared with other methods, our proposed method can generate natural inpainting results with reasonable semantics and richer textures with the help of the self-supervised Siamese inference network, the dense correspondence field, and the DAF module. It demonstrates that our proposed method is superior to the compared methods in terms of consistent structures and colors.

\subsection{Face Completion in the Wild}
Furthermore, we also conduct experiments on a real-world masked face dataset (RMFD) \cite{wang2020masked}. Note that there are no ground truth images in it. Therefore, we directly use our model and the compared models to evaluate on this dataset.  As shown in Fig. \ref{face completion results in the wild}, although there is a huge domain gap between our training sets and the real-world masked face dataset, our method can still generate relatively satisfactory results, which demonstrates the superiority of our proposed method. At the same time, some compared methods can not remove masks effectively, such as (d) and (e) in Fig. \ref{face completion results in the wild}. 

\textcolor{black}{We also provide the corresponding quantitative comparative experiments by using FID, Learned Perceptual Image Similarity (LPIPS) \cite{zhang2018unreasonable}, F1-Score and \textit{Realism} in Table \ref{quantitative resutls in the wild}. LPIPS measures the diversity of images by calculating the similarity in the feature space from the pre-trained AlexNet \cite{krizhevsky2012imagenet}. \textcolor{black}{F1-Score is the harmonic mean of \textit{recall} and \textit{precision}, where \textit{precision} is calculated by querying whether the each generated image is within the estimated manifold of real images and \textit{recall} is calculated by querying whether the each real image is within the estimated manifold of generated images \cite{kynkaanniemi2019improved}.} \textit{Realism} is a metric that reflects the distance between the image and the manifold: the closer the image is to the manifold, the higher \textit{Realism} is, and the further the image is from the manifold, the lower Realism is \cite{kynkaanniemi2019improved}. It clearly demonstrates the superiority of our proposed method in dealing with masked face images in real world.}

\subsection{Face Completion on Free-Form Occlusions}
\textcolor{black}{In the above three sections, we mainly conduct quantitative and qualitative experiments on face images with masks. In order to demonstrate the effectiveness of our method, we conduct experiments on the L2SFO dataset \cite{yuan2019face} in which face images are occluded by six common objects, i.e, masks, eyeglasses, sunglasses, cups, scarves, and hands. We conduct quantitative experiments on the testing set of L2SFO, and report the averaged results. we also retrain all the compared methods on the training sets of L2SFO for the sake of fairness. Table \ref{quantitative resutls on L2SFO} shows the performance of our proposed method against other compared methods. Our method outperforms all the other compared methods in three metrics on the testing sets as shown in this table. The results suggest that the proposed method can still extend to other kinds of occlusions.}

\begin{figure*}[h]
    \centering
    \includegraphics[scale=0.5]{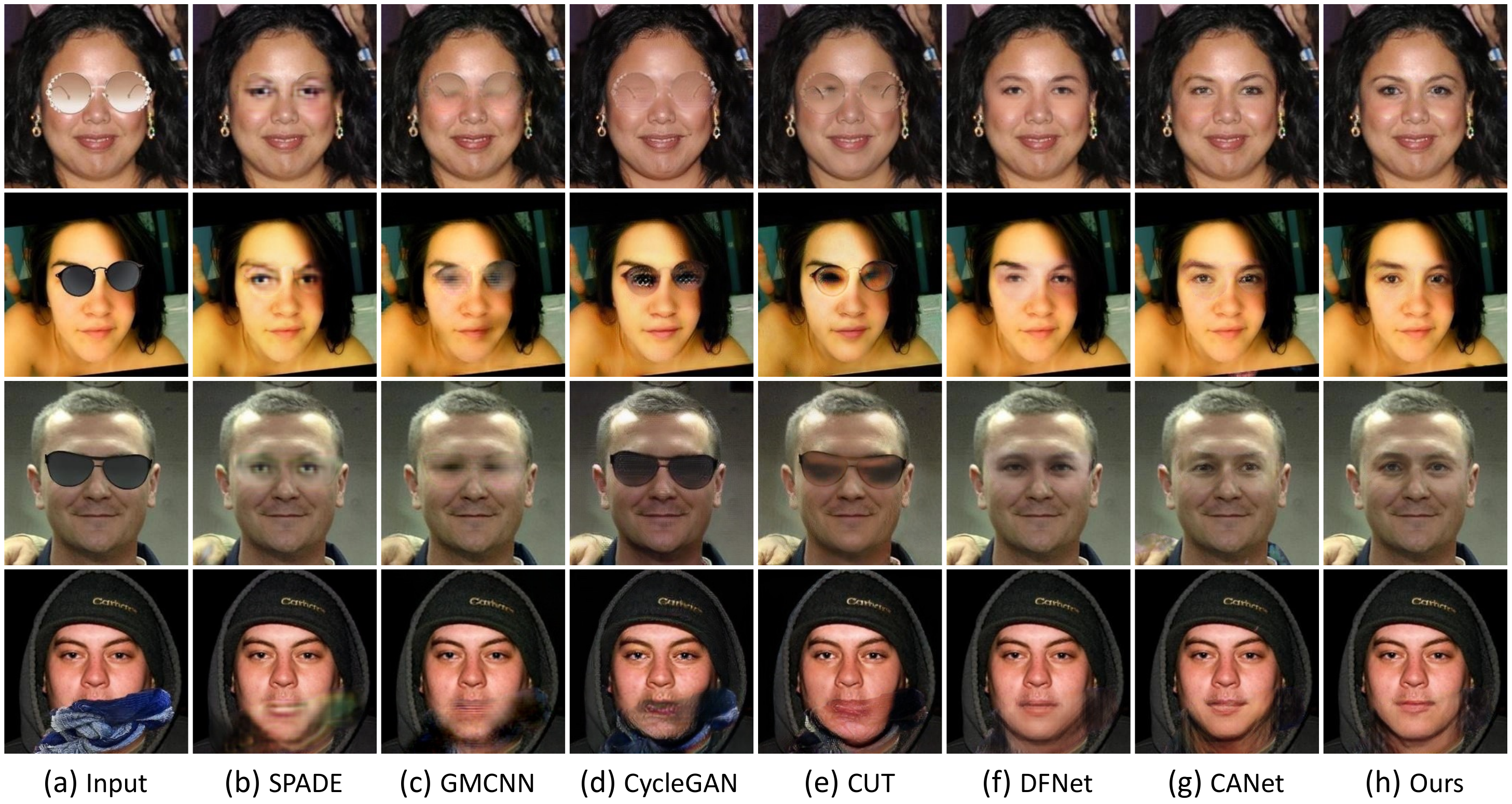}
    \caption{\textcolor{black}{Face completion results on the L2SFO dataset. From left to right, (a) are the input images. (b), (c), (d), (e), (f), (g) and (h) are the results generated by SPADE, GMCNN, CycleGAN, CUT, DFNet, CANet and ours method respectively.}}
    \label{face-completion-free-form}
\end{figure*}

\textcolor{black}{We also compare our proposed method with the state-of-the-art methods in terms of the visual quality on the testing set of L2SFO. As shown in Fig. \ref{face-completion-free-form}, we find that SPADE and GMCNN can remove occlusions, but there are serious artifacts in the generated images. CycleGAN and CUT fail to remove occlusions in some cases. Because they adopt unsupervised learning and hardly handle face images with complex occlusions. DFNet and CANet achieve relatively high-quality results. However, there are still artifacts in the generated face images produced by them. Different from all the compared methods, the proposed method can generate photo-realistic face images.}

\begin{figure}[ht]
    \centering
    \includegraphics[scale=0.5]{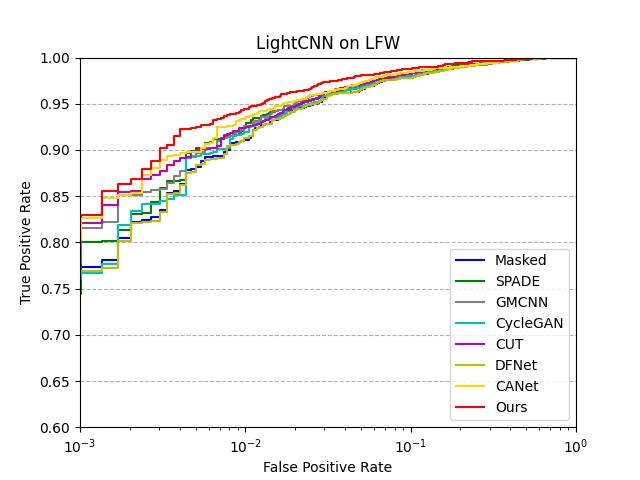}
    \caption{The ROC curves on the LFW dataset using LightCNN.}
    \label{ROC}
\end{figure}

\begin{table*}[]
\centering
\begin{tabular}{ccccccccc}
\hline
CL & \XSolidBrush      & \Checkmark      & \XSolidBrush      & \XSolidBrush      & \Checkmark      & \Checkmark      & \XSolidBrush      & \Checkmark      \\
DAF                  & \XSolidBrush      & \XSolidBrush      & \Checkmark      & \XSolidBrush      & \Checkmark      & \XSolidBrush      & \Checkmark      & \Checkmark      \\
UV map               & \XSolidBrush      & \XSolidBrush      & \XSolidBrush      & \Checkmark      & \XSolidBrush      & \Checkmark      & \Checkmark      & \Checkmark      \\ \hline
PSNR \ddag                & 29.71  & 31.04  & 31.82  & 31.11  & 32.02  & 32.20  & 32.40  & \textbf{32.82}  \\
SSIM \ddag                 & 0.9568 & 0.9664 & 0.9674 & 0.9663 & 0.9702 & 0.9708 & 0.9723 & \textbf{0.9755} \\
FID \dag                  & 1.9657 & 1.4729 & 1.3899 & 1.5059 & 1.259 & 1.3806 & 0.9872 & \textbf{0.9040} \\ \hline
\end{tabular}
\caption{\textcolor{black}{Ablation study experiments on the testing set of CelebA. \dag Lower is better. \ddag Higher is better. CL means Contrastive Learning.}}
\label{ablation study}
\end{table*}

\begin{figure*}[h]
    \centering
    \includegraphics[scale=0.5]{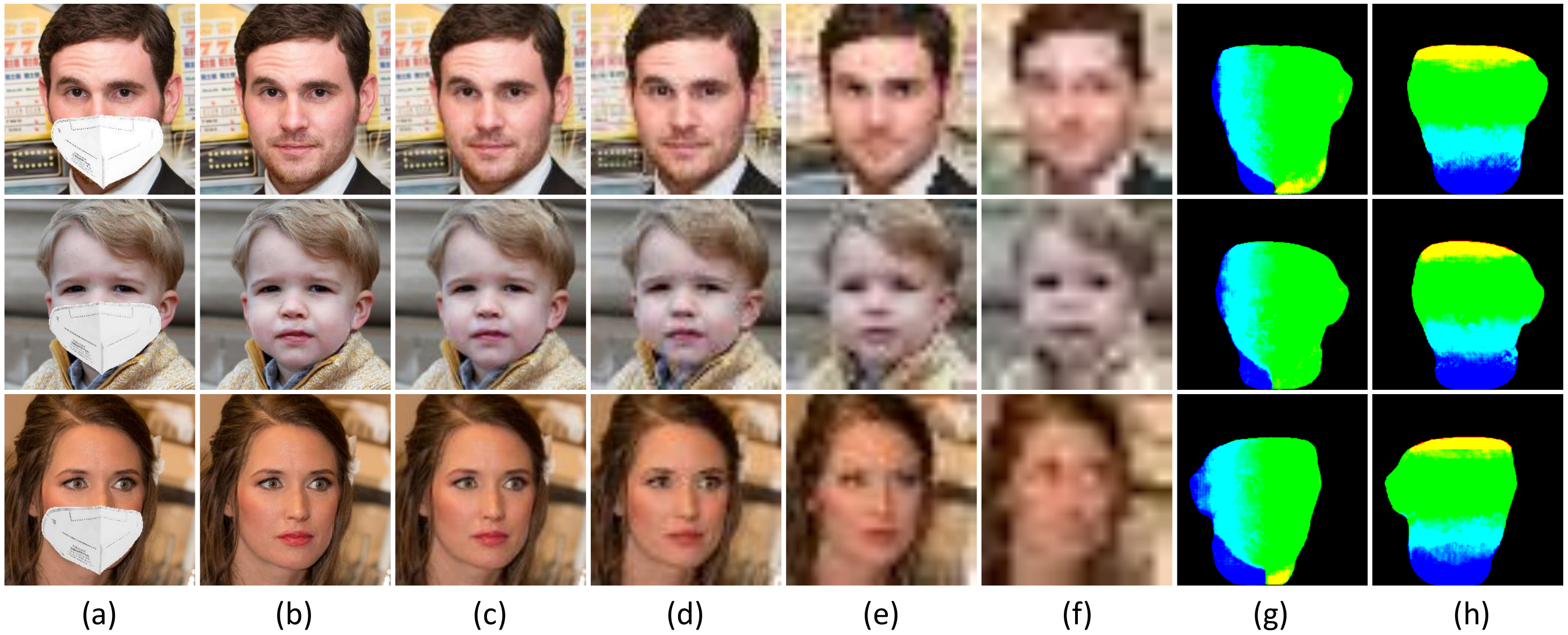}
    \caption{Images produced by the multi-scale decoder. (a) is the inputs with a 'KN95' mask. (b) is the final inpainting results. (c), (d), (e) and (f) are outputs at multi-scale. (g) and (h) are the estimated U and V maps, respectively.}
    \label{multi}
\end{figure*}

\begin{figure*}[h]
    \centering
    \includegraphics[scale=0.43]{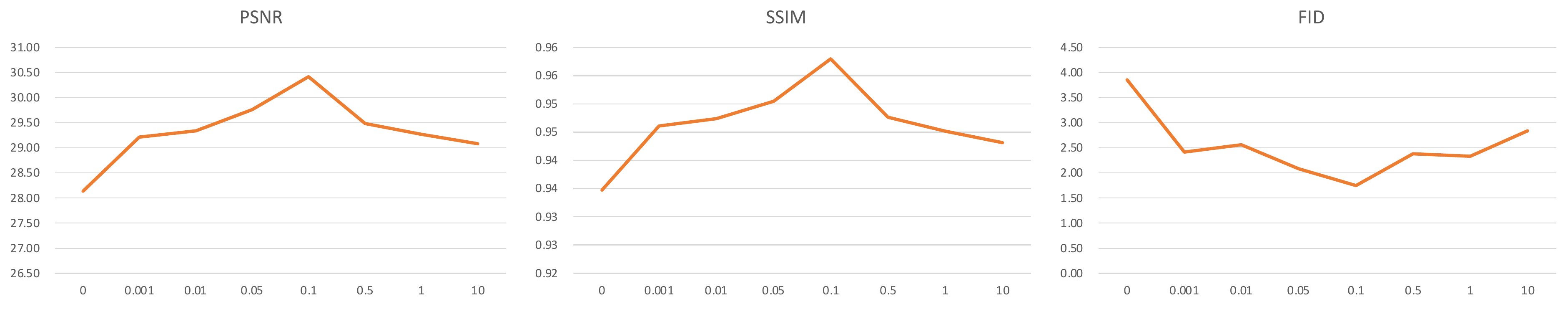}
    \caption{\textcolor{black}{Model performance affected by the weight of the UV loss on the FFHQ dataset.}}
    \label{loss_uv}
\end{figure*}

\begin{figure}[h]
    \centering
    \includegraphics[scale=0.5]{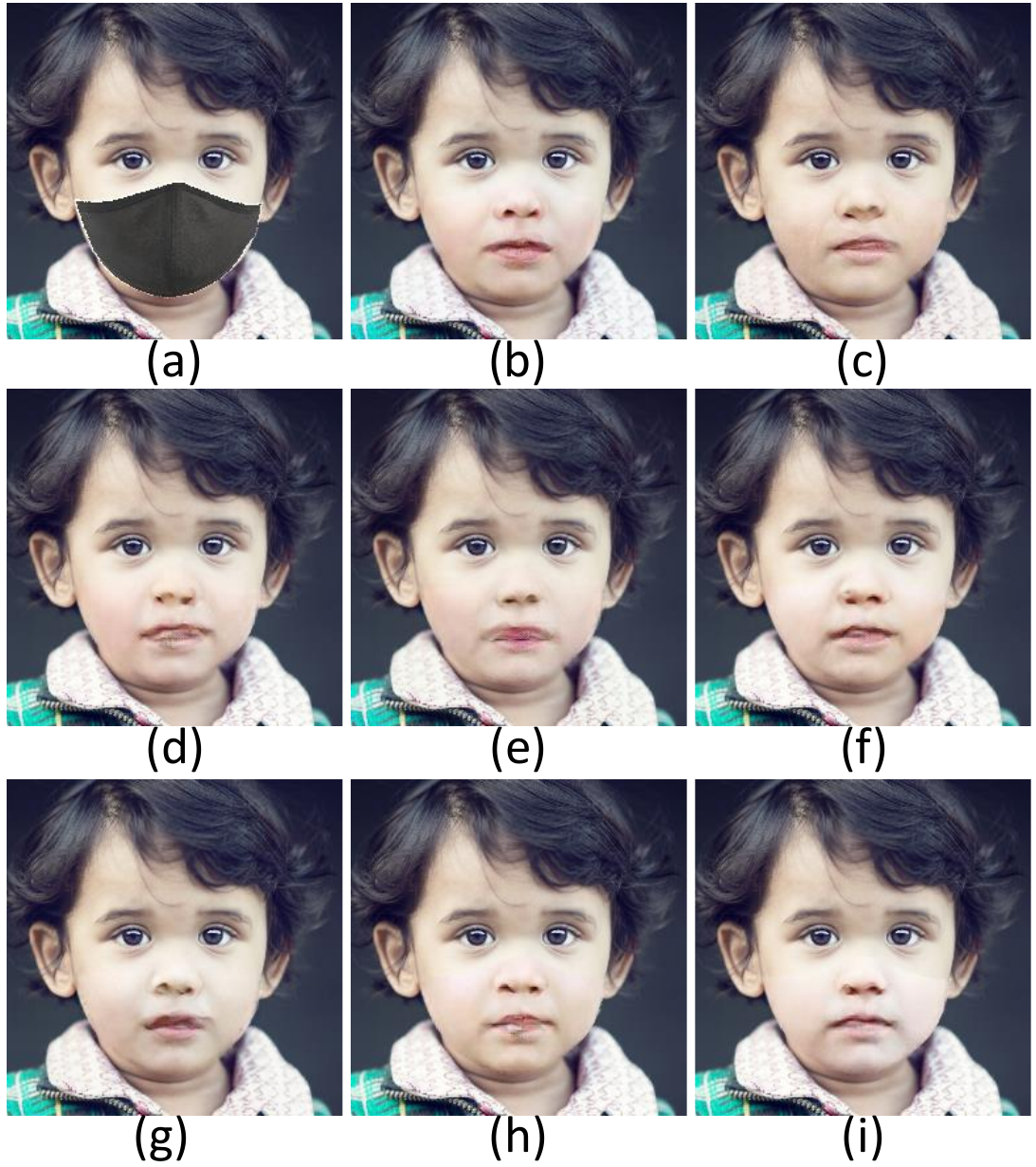}
    \caption{Images produced by the variants of our proposed method. (a) is the input with the 'cloth \#33333' mask. (i) is the result generated by the full model. (b)-(h) are results generated by the variant models according to Table \ref{ablation study}.}
    \label{ablation-fig}
\end{figure}

\subsection{Face Verification Results}
In order to quantitatively evaluate the feasibility of our method for face verification, we compare the results of our method and the compared methods on LFW \textcolor{black}{and IJB-C} following the testing protocol as described in Sec \ref{Datasets and Protocols}. Face verification experiments are conducted between the recovered probe set and the unchanged gallery set. Three publicly released face recognition models are tested: the LightCNN \cite{wu2018light}, ArcFace \cite{deng2019arcface} and FaceNet \cite{schroff2015facenet}. We use the area under the ROC curve (AUC), true positive rates at 1\% and 0.1\% (TPR@FPR=1\%, TPR@FPR=0.1\%) as the evaluation metrics in the experiments. The results are reported in Table \ref{face verification} \textcolor{black}{and Table \ref{face-verification-on-IJB-C}}.

We use the masked probe set as a baseline to demonstrate the influences of face completion on face verification. From Table \ref{face verification}, we can see that our method brings dramatic improvement to face verification. Because our method can keep geometric information intact and generate face images with consistent structures and colors. Compared with the baseline, our method can achieve an increase of more than 10\% in TPR@FPR=0.1\% \textcolor{black}{on LFW and an increase of 6.68\% in TPR@FPR=0.1\% on IJB-C}, which demonstrates that our proposed method can ameliorate the negative impact of masks. Similar to our method, the compared methods endeavor to recover face images. However, we find that the face verification performances of some compared methods decrease actually, especially in TPR@FPR=0.1\%. For instance, the performance of CycleGAN drops from 77.13\% to 76.06\% \textcolor{black}{on LFW}, a drop of about 1\% when taking the metric TPR@FAR=1\% and using LightCNN as the face feature extractor. \textcolor{black}{From Table \ref{face-verification-on-IJB-C}, we can also see that the compared methods do not show obvious advantages over the baseline ('Masked') on IJB-C. For example, the performance of CUT is 91.82\%, a very limited improvement of 0.006\% over the baseline when taking the metric TPR@FAR=1\% and using FaceNet as the face feature extractor.} \textcolor{black}{For the poor performances of compared methods on LFW and IJB-C,} the reason may lie in two aspects. The first reason is that the compared methods can not generate high-quality face images. The other reason is that they can not recover discriminative information of a face image due to the great negative effects of masks. We also present the ROC curves on LFW in Fig. \ref{ROC}. It is obvious that our method outperforms all the compared methods.

\subsection{Time Complexity}
We conduct the time complexity experiments on a single GPU (TITAN Xp) and CPU, respectively. To evaluate the inference time for different methods, we randomly sample 1,000 testing images and run forward one time for each image. Then we report the mean inference time for one image. As shown in Table \ref{the inference time on GPU}, our proposed method achieves a pleasing time performance compared with the other methods. It runs the second fast on a single TITAN Xp GPU. The fastest method is CUT on GPU. Because the number of parameters of CUT is only about a quarter of our method. However, as can be seen from Table \ref{quantitative results} and Fig. \ref{comaparison experments}, our method outperforms CUT with a large margin. When running on CPU, our proposed method is faster than SPADE, GMCNN, CycleGAN and CUT and achieves the comparable performance against DFNet.

\begin{table}[ht]
\setlength{\tabcolsep}{0.3mm}{
\begin{tabular}{ccccccc}
\hline
\begin{tabular}[c]{@{}c@{}}Inference\\ Time\end{tabular} & SPADE & GMCNN & CycleGAN & CUT   & DFNet & Ours  \\ \hline
GPU                                                      & 0.023 & 0.023 & 0.025    & 0.006 & 0.019 & 0.017 \\
CPU                                                      & 1.505 & 1.474 & 4.132    & 1.075 & 0.188 & 0.248 \\ \hline
\end{tabular}}
\caption{The inference time (seconds) on GPU and CPU.}
\label{the inference time on GPU}
\end{table}

\subsection{Ablation Study}
We investigate the effectiveness of different components of the proposed method on the testing set of CelebA. We train several variants of the proposed method: remove the self-supervised Siamese inference network (denote as contrastive learning), the DAF module, and/or the dense correspondence estimation (denoted as UV map). As shown in Table \ref{ablation study}, it clearly demonstrates that the self-supervised Siamese inference network, the DAF module, and the dense correspondence field estimation play important roles in determining the performance. As shown in Fig. \ref{ablation-fig}, the uncompleted models usually generate images with obvious artifacts, especially in boundaries while our full model can suppress color discrepancy and artifacts in boundaries and produce realistic inpainting results.

The multi-scale decoder can progressively refine the inpainting results at each scale. We also conduct experiments on the testing set of FFHQ. Then we visualize the images predicted by the decoder at several scales. As shown in Fig. \ref{multi}, it demonstrates that this multi-scale architecture is beneficial for decoding learned representations into generated images layer by layer. 

\textcolor{black}{We conduct sufficient experiments on the FFHQ dataset to explore the performance variation of our model affected by the weight of the UV loss function. We plot some figures according to the experimental results (Fig. 10). The horizontal axis represents the weight of the UV loss function. We use eight different weights to design the experiment, i.e, 0, 0.001, 0.01 0.05, 0.1, 0.5, 1 and 10. From Fig. \ref{loss_uv}, we can see that PSNR gradually increases with the increase of weight, reaches the maximum value when weight is equal to 0.1, and then drops sharply. The variation of SSIM is roughly the same as that of PSNR. The value of FID decreases dramatically from about 4 at the weight of 0 to around 2.5 at the weight of 0.001 and reaches the bottom (about 1.7) at the weight of 0.1. From these experiments, we can see that the UV loss (or \textit{Dense Correspondence Field Estimation}) plays an important role in determining the performance since it can keep the geometric information of the human face intact during the face completion process.
}

\section{Conclusion}
In this paper, we propose a novel two-stage paradigm image inpainting method to generate smoother results with reasonable semantics and richer textures. Specifically, the proposed method boosts the ability of the representation learning of the inference network by using contrastive learning. For keeping the geometric information of the input face image intact, we introduce a dense correspondence field that binds the 2D and 3D surface spaces into our network. We further design a novel dual attention fusion module, which can be embedded into decoder layers in a plug-and-play way. Extensive experiments show the superiority of our proposed method in generating smoother, more coherent, and fine-detailed results, and demonstrate our method can greatly improve the performance of face verification.

\section{Acknowledgments}
This work is partially funded by National Natural Science Foundation of China (Grant No. 62006228).

{\small
\bibliographystyle{ieee_fullname}
\bibliography{mybibfile}

\begin{thebibliography}{10}\itemsep=-1pt

\bibitem{alp2018densepose}
R{\i}za Alp~G{\"u}ler, Natalia Neverova, and Iasonas Kokkinos.
\newblock Densepose: Dense human pose estimation in the wild.
\newblock In {\em IEEE Conference on Computer Vision and Pattern Recognition},
  pages 7297--7306, 2018.

\bibitem{alp2017densereg}
Riza Alp~Guler, George Trigeorgis, Epameinondas Antonakos, Patrick Snape,
  Stefanos Zafeiriou, and Iasonas Kokkinos.
\newblock Densereg: Fully convolutional dense shape regression in-the-wild.
\newblock In {\em IEEE Conference on Computer Vision and Pattern Recognition},
  pages 6799--6808, 2017.

\bibitem{anwar2020masked}
Aqeel Anwar and Arijit Raychowdhury.
\newblock Masked face recognition for secure authentication.
\newblock {\em arXiv preprint arXiv:2008.11104}, 2020.

\bibitem{bachman2019learning}
Philip Bachman, R~Devon Hjelm, and William Buchwalter.
\newblock Learning representations by maximizing mutual information across
  views.
\newblock In {\em Advances in Neural Information Processing Systems}, pages
  15535--15545, 2019.

\bibitem{barnes2009patchmatch}
Connelly Barnes, Eli Shechtman, Adam Finkelstein, and Dan~B Goldman.
\newblock Patchmatch: A randomized correspondence algorithm for structural
  image editing.
\newblock {\em ACM Transactions on Graphics}, 28(3):24--33, 2009.

\bibitem{becker1992self}
Suzanna Becker and Geoffrey~E Hinton.
\newblock Self-organizing neural network that discovers surfaces in random-dot
  stereograms.
\newblock {\em Nature}, 355(6356):161--163, 1992.

\bibitem{bertalmio2000image}
Marcelo Bertalmio, Guillermo Sapiro, Vincent Caselles, and Coloma Ballester.
\newblock Image inpainting.
\newblock In {\em International Conference on Computer Graphics and Interactive
  Techniques}, pages 417--424, 2000.

\bibitem{blanz1999morphable}
Volker Blanz and Thomas Vetter.
\newblock A morphable model for the synthesis of 3d faces.
\newblock In {\em International Conference on Computer Graphics and Interactive
  Techniques}, pages 187--194, 1999.

\bibitem{booth2014optimal}
James Booth and Stefanos Zafeiriou.
\newblock Optimal uv spaces for facial morphable model construction.
\newblock In {\em International Conference on Image Processing}, pages
  4672--4676. IEEE, 2014.

\bibitem{cai2020semi}
Jiancheng Cai, Hu Han, Jiyun Cui, Jie Chen, Li Liu, and S~Kevin Zhou.
\newblock Semi-supervised natural face de-occlusion.
\newblock {\em IEEE Transactions on Information Forensics and Security},
  16:1044--1057, 2020.

\bibitem{cai2019fcsr}
Jiancheng Cai, Hu Han, Shiguang Shan, and Xilin Chen.
\newblock Fcsr-gan: Joint face completion and super-resolution via multi-task
  learning.
\newblock {\em IEEE Transactions on Biometrics, Behavior, and Identity
  Science}, 2(2):109--121, 2019.

\bibitem{cao2018learning}
Jie Cao, Yibo Hu, Hongwen Zhang, Ran He, and Zhenan Sun.
\newblock Learning a high fidelity pose invariant model for high-resolution
  face frontalization.
\newblock In {\em Advances in Neural Information Processing Systems}, pages
  2867--2877, 2018.

\bibitem{chen1982topological}
Lin Chen.
\newblock Topological structure in visual perception.
\newblock {\em Science}, 218(4573):699--700, 1982.

\bibitem{chen2022multi}
Rui Chen, Heng Zhang, and Jixin Liu.
\newblock Multi-attention augmented network for single image super-resolution.
\newblock {\em Pattern Recognition}, 122:108349, 2022.

\bibitem{chen2020SimCLR}
Ting Chen, Simon Kornblith, Mohammad Norouzi, and Geoffrey~E Hinton.
\newblock A simple framework for contrastive learning of visual
  representations.
\newblock In {\em International Conference on Machine Learning}, 2020.

\bibitem{chen2020dynamic}
Yinpeng Chen, Xiyang Dai, Mengchen Liu, Dongdong Chen, Lu Yuan, and Zicheng
  Liu.
\newblock Dynamic convolution: Attention over convolution kernels.
\newblock In {\em IEEE Conference on Computer Vision and Pattern Recognition},
  pages 11030--11039, 2020.

\bibitem{criminisi2004region}
Antonio Criminisi, Patrick P{\'e}rez, and Kentaro Toyama.
\newblock Region filling and object removal by exemplar-based image inpainting.
\newblock {\em IEEE Transactions on Image Processing}, 13(9):1200--1212, 2004.

\bibitem{deng2019arcface}
Jiankang Deng, Jia Guo, Niannan Xue, and Stefanos Zafeiriou.
\newblock Arcface: Additive angular margin loss for deep face recognition.
\newblock In {\em IEEE Conference on Computer Vision and Pattern Recognition},
  pages 4690--4699, 2019.

\bibitem{ding2018image}
Ding Ding, Sundaresh Ram, and Jeffrey~J Rodr{\'\i}guez.
\newblock Image inpainting using nonlocal texture matching and nonlinear
  filtering.
\newblock {\em IEEE Transactions on Image Processing}, 28(4):1705--1719, 2018.

\bibitem{ding2018perceptually}
Ding Ding, Sundaresh Ram, and Jeffrey~J Rodriguez.
\newblock Perceptually aware image inpainting.
\newblock {\em Pattern Recognition}, 83:174--184, 2018.

\bibitem{ding2021unsupervised}
Yuhe Ding, Xin Ma, Mandi Luo, Aihua Zheng, and Ran He.
\newblock Unsupervised contrastive photo-to-caricature translation based on
  auto-distortion.
\newblock In {\em International Conference on Pattern Recognition}, pages
  4520--4527. IEEE, 2021.

\bibitem{dosovitskiy2014discriminative}
Alexey Dosovitskiy, Jost~Tobias Springenberg, Martin Riedmiller, and Thomas
  Brox.
\newblock Discriminative unsupervised feature learning with convolutional
  neural networks.
\newblock {\em Advances in Neural Information Processing Systems}, 27:766--774,
  2014.

\bibitem{efros2001image}
Alexei~A Efros and William~T Freeman.
\newblock Image quilting for texture synthesis and transfer.
\newblock In {\em International Conference on Computer Graphics and Interactive
  Techniques}, pages 341--346, 2001.

\bibitem{goodfellow2014gan}
Ian Goodfellow, Jean Pouget-Abadie, Mehdi Mirza, Bing Xu, David Warde-Farley,
  Sherjil Ozair, Aaron Courville, and Yoshua Bengio.
\newblock Generative adversarial nets.
\newblock In {\em Advances in Neural Information Processing Systems}, pages
  2672--2680, 2014.

\bibitem{gross2010multi}
Ralph Gross, Iain Matthews, Jeffrey Cohn, Takeo Kanade, and Simon Baker.
\newblock Multi-pie.
\newblock {\em Image and Vision Computing}, 28(5):807--813, 2010.

\bibitem{hadsell2006dimensionality}
Raia Hadsell, Sumit Chopra, and Yann LeCun.
\newblock Dimensionality reduction by learning an invariant mapping.
\newblock In {\em IEEE Conference on Computer Vision and Pattern Recognition},
  volume~2, pages 1735--1742. IEEE, 2006.

\bibitem{he2020momentum}
Kaiming He, Haoqi Fan, Yuxin Wu, Saining Xie, and Ross Girshick.
\newblock Momentum contrast for unsupervised visual representation learning.
\newblock In {\em IEEE Conference on Computer Vision and Pattern Recognition},
  pages 9729--9738, 2020.

\bibitem{he2020non}
Wei He, Quanming Yao, Chao Li, Naoto Yokoya, Qibin Zhao, Hongyan Zhang, and
  Liangpei Zhang.
\newblock Non-local meets global: An integrated paradigm for hyperspectral
  image restoration.
\newblock {\em IEEE Transactions on Pattern Analysis and Machine Intelligence},
  2020.

\bibitem{hong2019dfnet}
Xin Hong, Pengfei Xiong, Renhe Ji, and Haoqiang Fan.
\newblock Deep fusion network for image completion.
\newblock In {\em ACM International Conference on Multimedia}, pages
  2033--2042, 2019.

\bibitem{hu2018senet}
Jie Hu, Li Shen, and Gang Sun.
\newblock Squeeze-and-excitation networks.
\newblock In {\em IEEE Conference on Computer Vision and Pattern Recognition},
  pages 7132--7141, 2018.

\bibitem{huang2008labeled}
Gary~B Huang, Marwan Mattar, Tamara Berg, and Eric Learned-Miller.
\newblock Labeled faces in the wild: A database forstudying face recognition in
  unconstrained environments.
\newblock 2008.

\bibitem{huang2019wavelet}
Huaibo Huang, Ran He, Zhenan Sun, and Tieniu Tan.
\newblock Wavelet domain generative adversarial network for multi-scale face
  hallucination.
\newblock {\em International Journal of Computer Vision}, 127(6-7):763--784,
  2019.

\bibitem{huang2017beyond}
Rui Huang, Shu Zhang, Tianyu Li, and Ran He.
\newblock Beyond face rotation: Global and local perception gan for
  photorealistic and identity preserving frontal view synthesis.
\newblock In {\em IEEE International Conference on Computer Vision}, pages
  2439--2448, 2017.

\bibitem{iizuka2017globalandlocal}
Satoshi Iizuka, Edgar Simo-Serra, and Hiroshi Ishikawa.
\newblock Globally and locally consistent image completion.
\newblock {\em ACM Transactions on Graphics}, 36(4):1--14, 2017.

\bibitem{jia2021inconsistency}
Gengyun Jia, Meisong Zheng, Chuanrui Hu, Xin Ma, Yuting Xu, Luoqi Liu, Yafeng
  Deng, and Ran He.
\newblock Inconsistency-aware wavelet dual-branch network for face forgery
  detection.
\newblock {\em IEEE Transactions on Biometrics, Behavior, and Identity
  Science}, 2021.

\bibitem{johnson2016perceptual}
Justin Johnson, Alexandre Alahi, and Li Fei-Fei.
\newblock Perceptual losses for real-time style transfer and super-resolution.
\newblock In {\em European Conference on Computer Vision}, pages 694--711.
  Springer, 2016.

\bibitem{karras2017progressive}
Tero Karras, Timo Aila, Samuli Laine, and Jaakko Lehtinen.
\newblock Progressive growing of gans for improved quality, stability, and
  variation.
\newblock In {\em International Conference on Learning Representations}, 2018.

\bibitem{karras2019style}
Tero Karras, Samuli Laine, and Timo Aila.
\newblock A style-based generator architecture for generative adversarial
  networks.
\newblock In {\em IEEE Conference on Computer Vision and Pattern Recognition},
  pages 4401--4410, 2019.

\bibitem{kim2019recurrent}
Dahun Kim, Sanghyun Woo, Joon-Young Lee, and In~So Kweon.
\newblock Recurrent temporal aggregation framework for deep video inpainting.
\newblock {\em IEEE Transactions on Pattern Analysis and Machine Intelligence},
  42(5):1038--1052, 2019.

\bibitem{krizhevsky2012imagenet}
Alex Krizhevsky, Ilya Sutskever, and Geoffrey~E Hinton.
\newblock Imagenet classification with deep convolutional neural networks.
\newblock {\em Advances in Neural Information Processing Systems},
  25:1097--1105, 2012.

\bibitem{kynkaanniemi2019improved}
Tuomas Kynk{\"a}{\"a}nniemi, Tero Karras, Samuli Laine, Jaakko Lehtinen, and
  Timo Aila.
\newblock Improved precision and recall metric for assessing generative models.
\newblock {\em Advances in Neural Information Processing Systems}, 2019.

\bibitem{li2019disentangled}
Yi Li, Huaibo Huang, Jie Cao, Ran He, and Tieniu Tan.
\newblock Disentangled representation learning of makeup portraits in the wild.
\newblock {\em International Journal of Computer Vision}, pages 1--19, 2019.

\bibitem{li2020learning}
Zhihang Li, Yibo Hu, Ran He, and Zhenan Sun.
\newblock Learning disentangling and fusing networks for face completion under
  structured occlusions.
\newblock {\em Pattern Recognition}, 99:107073, 2020.

\bibitem{liu2018image}
Guilin Liu, Fitsum~A Reda, Kevin~J Shih, Ting-Chun Wang, Andrew Tao, and Bryan
  Catanzaro.
\newblock Image inpainting for irregular holes using partial convolutions.
\newblock In {\em European Conference on Computer Vision}, pages 85--100, 2018.

\bibitem{liu2015faceattributes}
Ziwei Liu, Ping Luo, Xiaogang Wang, and Xiaoou Tang.
\newblock Deep learning face attributes in the wild.
\newblock In {\em IEEE International Conference on Computer Vision}, December
  2015.

\bibitem{luo2021fa}
Mandi Luo, Jie Cao, Xin Ma, Xiaoyu Zhang, and Ran He.
\newblock Fa-gan: Face augmentation gan for deformation-invariant face
  recognition.
\newblock {\em IEEE Transactions on Information Forensics and Security},
  16:2341--2355, 2021.

\bibitem{luo2021partial}
Mandi Luo, Xin Ma, Zhihang Li, Jie Cao, and Ran He.
\newblock Partial nir-vis heterogeneous face recognition with automatic
  saliency search.
\newblock {\em IEEE Transactions on Information Forensics and Security}, 2021.

\bibitem{ma2020free}
Xin Ma, Xiaoqiang Zhou, Huaibo Huang, Zhenhua Chai, Xiaolin Wei, and Ran He.
\newblock Free-form image inpainting via contrastive attention network.
\newblock In {\em International Conference on Pattern Recognition}, 2020.

\bibitem{mustikovela2020self-viewpoint}
Siva~Karthik Mustikovela, Varun Jampani, Shalini~De Mello, Sifei Liu, Umar
  Iqbal, Carsten Rother, and Jan Kautz.
\newblock Self-supervised viewpoint learning from image collections.
\newblock In {\em IEEE Conference on Computer Vision and Pattern Recognition},
  pages 3971--3981, 2020.

\bibitem{Nazeri_2019_ICCV}
Kamyar Nazeri, Eric Ng, Tony Joseph, Faisal Qureshi, and Mehran Ebrahimi.
\newblock Edgeconnect: Structure guided image inpainting using edge prediction.
\newblock In {\em IEEE International Conference on Computer Vision Workshops},
  Oct 2019.

\bibitem{obeso2021visual}
Abraham~Montoya Obeso, Jenny Benois-Pineau, Mireya Sara{\'\i}~Garc{\'\i}a
  V{\'a}zquez, and Alejandro {\'A}lvaro~Ram{\'\i}rez Acosta.
\newblock Visual vs internal attention mechanisms in deep neural networks for
  image classification and object detection.
\newblock {\em Pattern Recognition}, page 108411, 2021.

\bibitem{park2020contrastive}
Taesung Park, Alexei~A Efros, Richard Zhang, and Jun-Yan Zhu.
\newblock Contrastive learning for unpaired image-to-image translation.
\newblock In {\em European Conference on Computer Vision}, pages 319--345.
  Springer, 2020.

\bibitem{park2019semantic}
Taesung Park, Ming-Yu Liu, Ting-Chun Wang, and Jun-Yan Zhu.
\newblock Semantic image synthesis with spatially-adaptive normalization.
\newblock In {\em IEEE Conference on Computer Vision and Pattern Recognition},
  pages 2337--2346, 2019.

\bibitem{pathak2016contextencoder}
Deepak Pathak, Philipp Krahenbuhl, Jeff Donahue, Trevor Darrell, and Alexei~A
  Efros.
\newblock Context encoders: Feature learning by inpainting.
\newblock In {\em IEEE Conference on Computer Vision and Pattern Recognition},
  pages 2536--2544, 2016.

\bibitem{paysan20093d}
Pascal Paysan, Reinhard Knothe, Brian Amberg, Sami Romdhani, and Thomas Vetter.
\newblock A 3d face model for pose and illumination invariant face recognition.
\newblock In {\em IEEE International Conference on Advanced Video and Signal
  Based Surveillance}, pages 296--301. IEEE, 2009.

\bibitem{pei2021all}
Zhao Pei, Min Jin, Yanning Zhang, Miao Ma, and Yee-Hong Yang.
\newblock All-in-focus synthetic aperture imaging using generative adversarial
  network-based semantic inpainting.
\newblock {\em Pattern Recognition}, 111:107669, 2021.

\bibitem{perez2003poisson}
Patrick P{\'e}rez, Michel Gangnet, and Andrew Blake.
\newblock Poisson image editing.
\newblock In {\em ACM SIGGRAPH 2003 Papers}, pages 313--318. 2003.

\bibitem{ren2019structureflow}
Yurui Ren, Xiaoming Yu, Ruonan Zhang, Thomas~H Li, Shan Liu, and Ge Li.
\newblock Structureflow: Image inpainting via structure-aware appearance flow.
\newblock In {\em IEEE International Conference on Computer Vision}, pages
  181--190, 2019.

\bibitem{roth2015unconstrained}
Joseph Roth, Yiying Tong, and Xiaoming Liu.
\newblock Unconstrained 3d face reconstruction.
\newblock In {\em IEEE Conference on Computer Vision and Pattern Recognition},
  pages 2606--2615, 2015.

\bibitem{schroff2015facenet}
Florian Schroff, Dmitry Kalenichenko, and James Philbin.
\newblock Facenet: A unified embedding for face recognition and clustering.
\newblock In {\em IEEE Conference on Computer Vision and Pattern Recognition},
  pages 815--823, 2015.

\bibitem{tu2019joint}
Xiaoguang Tu, Jian Zhao, Zihang Jiang, Yao Luo, Mei Xie, Yang Zhao, Linxiao He,
  Zheng Ma, and Jiashi Feng.
\newblock Joint 3d face reconstruction and dense face alignment from a single
  image with 2d-assisted self-supervised learning.
\newblock {\em arXiv preprint arXiv:1903.09359}, 2019.

\bibitem{tu20203d}
Xiaoguang Tu, Jian Zhao, Mei Xie, Zihang Jiang, Akshaya Balamurugan, Yao Luo,
  Yang Zhao, Lingxiao He, Zheng Ma, and Jiashi Feng.
\newblock 3d face reconstruction from a single image assisted by 2d face images
  in the wild.
\newblock {\em IEEE Transactions on Multimedia}, 2020.

\bibitem{wang2020multistage}
Ning Wang, Sihan Ma, Jingyuan Li, Yipeng Zhang, and Lefei Zhang.
\newblock Multistage attention network for image inpainting.
\newblock {\em Pattern Recognition}, 106:107448, 2020.

\bibitem{wang2019laplacian}
Qiang Wang, Huijie Fan, Gan Sun, Yang Cong, and Yandong Tang.
\newblock Laplacian pyramid adversarial network for face completion.
\newblock {\em Pattern Recognition}, 88:493--505, 2019.

\bibitem{wang2018non-local}
Xiaolong Wang, Ross Girshick, Abhinav Gupta, and Kaiming He.
\newblock Non-local neural networks.
\newblock In {\em IEEE Conference on Computer Vision and Pattern Recognition},
  pages 7794--7803, 2018.

\bibitem{wang2015unsupervised}
Xiaolong Wang and Abhinav Gupta.
\newblock Unsupervised learning of visual representations using videos.
\newblock In {\em IEEE Conference on Computer Vision and Pattern Recognition},
  pages 2794--2802, 2015.

\bibitem{wang2018image}
Yi Wang, Xin Tao, Xiaojuan Qi, Xiaoyong Shen, and Jiaya Jia.
\newblock Image inpainting via generative multi-column convolutional neural
  networks.
\newblock In {\em Advances in Neural Information Processing Systems}, pages
  331--340, 2018.

\bibitem{wang2020masked}
Zhongyuan Wang, Guangcheng Wang, Baojin Huang, Zhangyang Xiong, Qi Hong, Hao
  Wu, Peng Yi, Kui Jiang, Nanxi Wang, Yingjiao Pei, et~al.
\newblock Masked face recognition dataset and application.
\newblock {\em arXiv preprint arXiv:2003.09093}, 2020.

\bibitem{Whitelam2017IARPAJB}
Cameron Whitelam, Emma Taborsky, Austin Blanton, Brianna Maze, Jocelyn~C.
  Adams, Tim Miller, Nathan~D. Kalka, Anil~K. Jain, James~A. Duncan, Kristen~E
  Allen, Jordan Cheney, and Patrick Grother.
\newblock Iarpa janus benchmark-b face dataset.
\newblock {\em IEEE Conference on Computer Vision and Pattern Recognition
  Workshops}, pages 592--600, 2017.

\bibitem{wu2018light}
Xiang Wu, Ran He, Zhenan Sun, and Tieniu Tan.
\newblock A light cnn for deep face representation with noisy labels.
\newblock {\em IEEE Transactions on Information Forensics and Security},
  13(11):2884--2896, 2018.

\bibitem{wu2018unsupervised}
Zhirong Wu, Yuanjun Xiong, Stella~X Yu, and Dahua Lin.
\newblock Unsupervised feature learning via non-parametric instance
  discrimination.
\newblock In {\em IEEE Conference on Computer Vision and Pattern Recognition},
  pages 3733--3742, 2018.

\bibitem{xie2019image}
Chaohao Xie, Shaohui Liu, Chao Li, Ming-Ming Cheng, Wangmeng Zuo, Xiao Liu,
  Shilei Wen, and Errui Ding.
\newblock Image inpainting with learnable bidirectional attention maps.
\newblock In {\em IEEE International Conference on Computer Vision}, pages
  8858--8867, 2019.

\bibitem{xu2010image}
Zongben Xu and Jian Sun.
\newblock Image inpainting by patch propagation using patch sparsity.
\newblock {\em IEEE Transactions on Image Processing}, 19(5):1153--1165, 2010.

\bibitem{yu2018generative}
Jiahui Yu, Zhe Lin, Jimei Yang, Xiaohui Shen, Xin Lu, and Thomas~S Huang.
\newblock Generative image inpainting with contextual attention.
\newblock In {\em IEEE Conference on Computer Vision and Pattern Recognition},
  pages 5505--5514, 2018.

\bibitem{yu2018contextualattention}
Jiahui Yu, Zhe Lin, Jimei Yang, Xiaohui Shen, Xin Lu, and Thomas~S Huang.
\newblock Generative image inpainting with contextual attention.
\newblock In {\em IEEE Conference on Computer Vision and Pattern Recognition},
  pages 5505--5514, 2018.

\bibitem{yu2019free}
Jiahui Yu, Zhe Lin, Jimei Yang, Xiaohui Shen, Xin Lu, and Thomas~S Huang.
\newblock Free-form image inpainting with gated convolution.
\newblock In {\em IEEE International Conference on Computer Vision}, pages
  4471--4480, 2019.

\bibitem{yuan2019face}
Xiaowei Yuan and In~Kyu Park.
\newblock Face de-occlusion using 3d morphable model and generative adversarial
  network.
\newblock In {\em IEEE International Conference on Computer Vision}, pages
  10062--10071, 2019.

\bibitem{zeng2021feature}
Yuan Zeng, Yi Gong, and Jin Zhang.
\newblock Feature learning and patch matching for diverse image inpainting.
\newblock {\em Pattern Recognition}, page 108036, 2021.

\bibitem{zeng2020high}
Yu Zeng, Zhe Lin, Jimei Yang, Jianming Zhang, Eli Shechtman, and Huchuan Lu.
\newblock High-resolution image inpainting with iterative confidence feedback
  and guided upsampling.
\newblock In {\em European Conference on Computer Vision}, 2020.

\bibitem{zhan2020self}
Xiaohang Zhan, Xingang Pan, Bo Dai, Ziwei Liu, Dahua Lin, and Chen~Change Loy.
\newblock Self-supervised scene de-occlusion.
\newblock In {\em IEEE Conference on Computer Vision and Pattern Recognition},
  June 2020.

\bibitem{zhan2020self-occlusion}
Xiaohang Zhan, Xingang Pan, Bo Dai, Ziwei Liu, Dahua Lin, and Chen~Change Loy.
\newblock Self-supervised scene de-occlusion.
\newblock In {\em IEEE Conference on Computer Vision and Pattern Recognition},
  pages 3784--3792, 2020.

\bibitem{zhang2018unreasonable}
Richard Zhang, Phillip Isola, Alexei~A Efros, Eli Shechtman, and Oliver Wang.
\newblock The unreasonable effectiveness of deep features as a perceptual
  metric.
\newblock In {\em IEEE Conference on Computer Vision and Pattern Recognition},
  pages 586--595, 2018.

\bibitem{zhang2017demeshnet}
Shu Zhang, Ran He, Zhenan Sun, and Tieniu Tan.
\newblock Demeshnet: Blind face inpainting for deep meshface verification.
\newblock {\em IEEE Transactions on Information Forensics and Security},
  13(3):637--647, 2017.

\bibitem{zhang2018image}
Yulun Zhang, Kunpeng Li, Kai Li, Lichen Wang, Bineng Zhong, and Yun Fu.
\newblock Image super-resolution using very deep residual channel attention
  networks.
\newblock In {\em European Conference on Computer Vision}, pages 286--301,
  2018.

\bibitem{zhou2020learning}
Tong Zhou, Changxing Ding, Shaowen Lin, Xinchao Wang, and Dacheng Tao.
\newblock Learning oracle attention for high-fidelity face completion.
\newblock In {\em IEEE Conference on Computer Vision and Pattern Recognition},
  pages 7680--7689, 2020.

\bibitem{zhu2017unpaired}
Jun-Yan Zhu, Taesung Park, Phillip Isola, and Alexei~A Efros.
\newblock Unpaired image-to-image translation using cycle-consistent
  adversarial networks.
\newblock In {\em IEEE International Conference on Computer Vision}, pages
  2223--2232, 2017.

\bibitem{zhu2016face}
Xiangyu Zhu, Zhen Lei, Xiaoming Liu, Hailin Shi, and Stan~Z Li.
\newblock Face alignment across large poses: A 3d solution.
\newblock In {\em IEEE Conference on Computer Vision and Pattern Recognition},
  pages 146--155, 2016.

\bibitem{zhuang2019local}
Chengxu Zhuang, Alex~Lin Zhai, and Daniel Yamins.
\newblock Local aggregation for unsupervised learning of visual embeddings.
\newblock In {\em IEEE International Conference on Computer Vision}, pages
  6002--6012, 2019.

\end{thebibliography}
}

\end{document}